\pdfoutput=1
\documentclass[11pt]{article}

\usepackage[]{acl}
\definecolor{darkred}{rgb}{0.8, 0.0, 0.0}
\usepackage{times}
\usepackage{latexsym}
\usepackage{hyperref}
\usepackage{booktabs}
\usepackage{multirow}
\usepackage{amsmath}
\usepackage{seqsplit}
\usepackage{amsmath}
\usepackage{amssymb}
\usepackage{amsfonts}
\usepackage{graphicx}
\usepackage{subcaption}
\usepackage{longtable}
\usepackage{float}

\usepackage[T1]{fontenc}
\usepackage[utf8]{inputenc}

\usepackage{microtype}

\usepackage{inconsolata}

\usepackage{graphicx}
\usepackage{color}
\usepackage{xcolor, colortbl}
\usepackage[most]{tcolorbox}
\usepackage{soul}

\definecolor{lightpink}{rgb}{0.945, 0.816, 0.804}
\definecolor{lightgreen}{rgb}{0.851, 0.906, 0.839}
\definecolor{lightblue}{rgb}{0.8, 0.9, 1}
\definecolor{lightyellow}{rgb}{0.992, 0.949, 0.816}

\newtcolorbox[list inside=prompt,auto counter,number within=section]{prompt}[1][]{
    colbacktitle=black!60,
    coltitle=white,
    fontupper=\footnotesize,
    boxsep=5pt,
    breakable,
    enhanced,
    left=0pt,
    right=0pt,
    top=0pt,
    bottom=0pt,
    boxrule=1pt,
    #1   
}
\title{Large Language Models Penetration in Scholarly Writing and Peer Review}

\newcommand{\cuhksz}{$^1$}
\newcommand{\uestc}{$^2$}
\newcommand{\whut}{$^3$}
\newcommand{\ku}{$^4$}
\newcommand{\sribd}{$^5$}

\author{
 \textbf{Li Zhou\cuhksz},
 \textbf{Ruijie Zhang\uestc},
 \textbf{Xunlian Dai\whut},
 \textbf{Daniel Hershcovich\ku},
 \textbf{Haizhou Li\cuhksz$^,$\sribd}
\\
{\cuhksz}The Chinese University of Hong Kong, Shenzhen \\ 
{\uestc}University of Electronic Science and Technology of China \\ 
{\whut}Wuhan University of Technology {\ku}University of Copenhagen \\
{\sribd}Shenzhen Research Institute of Big Data\\
 \small{
   \texttt{lizhou21@cuhk.edu.cn}
 }
}

\begin{document}
\maketitle
\begin{abstract}

While the widespread use of Large Language Models (LLMs) brings convenience, it also raises concerns about the credibility of academic research and scholarly processes. To better understand these dynamics, we evaluate the penetration of LLMs across academic workflows from multiple perspectives and dimensions, providing compelling evidence of their growing influence. We propose a framework with two components: \texttt{ScholarLens}, a curated dataset of human-written and LLM-generated content across scholarly writing and peer review for multi-perspective evaluation, and \texttt{LLMetrica}, a tool for assessing LLM penetration using rule-based metrics and model-based detectors for multi-dimensional evaluation. Our experiments demonstrate the effectiveness of \texttt{LLMetrica}, revealing the increasing role of LLMs in scholarly processes. These findings emphasize the need for transparency, accountability, and ethical practices in LLM usage to maintain academic credibility.

\end{abstract}

\section{Introduction}

The rapid advancement of large language models ~\cite[LLMs;][] {achiam2023gpt, dubey2024llama, guo2025deepseek} is significantly shaping the scholarly landscape~\cite{hosseini2023fighting, geng2024impact}.
These technologies assist in various stages of scholarly work, from brainstorming and overcoming ``blank-sheet syndrome''~\cite{altmae2023artificial, baek2024researchagent, eger2025transformingsciencelargelanguage} to supporting paper writing~\cite{wang-etal-2018-paper, birhane2023science, khalifa2024using, rehman2025can}. 
More recently, LLMs have also been considered as tools in the peer review process~\cite{du-etal-2024-llms, zhou-etal-2024-llm, yu2024your, jin-etal-2024-agentreview, zou2024chatgpt}.

However, LLM-generated content often reflects lower quality and inherent biases~\cite{brooks-etal-2024-rise, du-etal-2024-llms}, such as factual inconsistencies~\cite{yang-etal-2024-fizz, chuang-etal-2024-lookback} and hallucinations~\cite{tang2024llms, chuang-etal-2024-lookback}. Research also indicates that LLMs in the review process can lead to higher paper acceptance rates~\cite{latona2024ai, jin-etal-2024-agentreview} and tend to use more positive language than human reviewers~\cite{zhou-etal-2024-llm}. 
This raises concerns about the rigor of scientific research~\cite{sun2024metawriter}, highlighting the importance of ensuring transparency and accountability in the use of LLMs within academic workflows~\cite{lund2024can}.
% This raises concerns about the rigor of scientific research~\cite{sun2024metawriter}. 
% Given these developments, ensuring the transparent and accountable use of LLMs in academic workflows is crucial.

% and understanding the extent and trends of their penetration into scholarly processes is becoming increasingly important.

\begin{figure*}[t]
    \centering
    \includegraphics[width=1.0\linewidth]{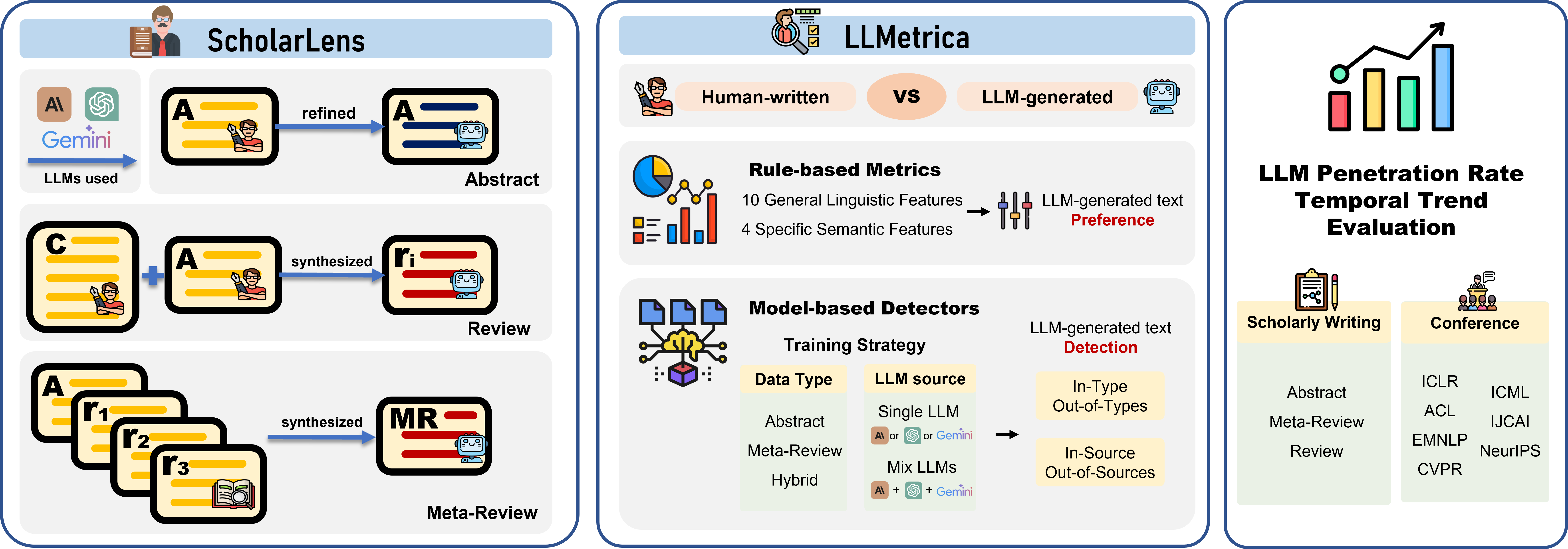}
    \caption{Pipeline Overview of Our Work: (1) \textbf{\texttt{ScholarLens} Curation} (\S\ref{sec: ScholarLens}): A designed dataset used to evaluate the effectiveness of metrics and train detection models; (2) \textbf{\texttt{LLMetrica} framework} (\S\ref{sec: LLMetria}): The proposed method for distinguishing human-written from LLM-generated texts; (3) \textbf{Experiments} (\S\ref{sec:Experiments}): Evaluating the effectiveness of \texttt{LLMetrica} and applying it to real-world data to assess LLM penetration rates in scholarly writing and peer process. \textit{Symbolically}, $\mathrm{P}=\left\{ \mathrm{T},\mathrm{A},\mathrm{C},\mathrm{R},\mathrm{MR} \right\}$ represents a research paper, where $\mathrm{T}$, $\mathrm{A}$ and $\mathrm{C}$ denote its title, abstract and main content, $\mathrm{R}=\left\{ r_i \right\}$ represents the individual reviews, and $\mathrm{MR}$ denotes the meta-review.}
    \label{fig:pipeline}
\end{figure*}

To address these concerns, we propose a comprehensive evaluation framework aimed at revealing the increasing penetration of LLMs in scholarly writing and peer review. 
This framework takes a multi-perspective view by integrating diverse scholarly data types and employs a multi-dimensional methodology that utilizes a range of evaluation methods to provide a reliable and nuanced understanding of LLM usage trends. Figure~\ref{fig:pipeline} illustrate the pipeline of our work, and our \textbf{contributions} are as follows:
% \footnote{Dataset and code are available in the final version.}

\begin{itemize}
    \item We introduce \texttt{ScholarLens}, a curated dataset for developing technical measurement methods, comprising both human-written and LLM-generated content (\S\ref{sec: ScholarLens}).
    \item We propose \texttt{LLMetrica}, a tool for assessing LLM penetration in scholarly workflows, combining rule-based metrics to analyze linguistic and semantic features with model-based detectors to identify LLM-generated content (\S\ref{sec: LLMetria}).
    \item Our experiments demonstrate the effectiveness of \texttt{LLMetrica}, consistently showing the increasing penetration of LLMs in scholarly writing and peer review from multiple perspectives and dimensions (\S\ref{sec:Experiments}).
    % \item Our experiments demonstrate the effectiveness of \texttt{LLMetrica}, consistently showing the increasing penetration of LLMs in scholarly writing and peer review of real-word data on real-world data from multiple perspectives and dimensions (\S\ref{sec:Experiments}).
\end{itemize}
% Our work reveal the growing penetration of LLMs in scholarly processes,
Our findings emphasize the need for transparency, accountability, and ethical practices in LLM usage to maintain the credibility of academic research.

\section{Related Work}

% \lz{Li is writing this section}

Recent discussions in the scientific community have focused on improving the peer review process~\cite{gurevych_et_al:DagRep.14.1.130, kuznetsov2024natural} to address issues like misalignment between reviewers and paper topics, as well as social~\cite{huber2022nobel, tomkins2017reviewer, manzoor2021uncovering} and cognitive biases~\cite{lee2015commensuration, stelmakh2021prior}. Proposed solutions include enhancing structural incentives for reviewers~\cite{rogers-augenstein-2020-improve}, using natural language processing for intelligent support~\cite{kuznetsov-etal-2022-revise, zyska-etal-2023-care, dycke-etal-2023-nlpeer, guo-etal-2023-automatic, kumar-etal-2023-reviewers}, and other policy recommendations~\cite{dycke-etal-2022-yes}. Furthermore, some studies focus on the collection and analysis of review data~\cite{kennard-etal-2022-disapere, staudinger-etal-2024-analysis, darcy-etal-2024-aries}.\footnote{\href{https://arr-data.aclweb.org/}{ACL Rolling Review Data Collection (ARR-DC)}.} 
However, these efforts largely focus on human reviewers: what if instead the reviewers are LLMs~\cite{weber2024other, gao2024reviewer2, hossain2025llmsmetareviewersassistantscase}?

Previous research has demonstrated that human-written and LLM-generated texts exhibit distinct linguistic characteristics~\cite{cheng2024beyond, song-etal-2025-assessing}.
For instance, LLM-generated texts often display the recurrent use of specific syntactic templates~\cite{shaib-etal-2024-detection}, which largely reflect patterns learned from the training data and highlight the model's memorization capacity~\cite{karamolegkou-etal-2023-copyright, zeng-etal-2024-exploring, zhu-etal-2024-beyond}.
Furthermore, some studies develop detection models to identify LLM-generated text~\cite{antoun-etal-2024-text, cheng2024beyond, xu-etal-2024-detecting, kumar-etal-2024-quis, abassy-etal-2024-llm}.
However, \citet{cheng2024beyond} points out that supervised detectors exhibit poor cross-domain generalizability.

Unlike previous work, we simulate LLM usage in scholarly writing and peer review, creating a comparison between LLM-generated and human-written texts.
We also develop a robust framework to assess the distinctive tendencies of LLM-generated content and identify it within the scholarly domain. This framework offers a comprehensive approach to evaluating LLM penetration from multiple perspectives and dimensions.

\section{\texttt{ScholarLens} Curation}
\label{sec: ScholarLens}
% To investigate the penetration rate of LLMs in these stages, 

In this section, we detail the process of curating \texttt{ScholarLens}, including the consideration of data types and the setup for collecting both human-written and LLM-generated text.

% the collection setup of human-written and LLM-generated text.

% criteria for paper selection, the collection of human-written and LLM-generated reviews, the annotation procedure, and the measures taken to ensure data quality.

% \subsection{What Aspects of LLMs in Academia Are Considered?}

% \subsection{How to Construct LLM-Generated Data for Comparison?}

% \begin{table}[]
% \centering
% \scalebox{0.85}{
% \begin{tabular}{@{}l|rrrrrr@{}}
% \toprule
% \textbf{Source} & \textbf{$\leq 2019$} & \textbf{$2020$} & \textbf{$2021$} & \textbf{$2022$} & \textbf{$2023$} & \textbf{$2024$} \\ \midrule
% \textbf{ICLR}   & 2831           & 2213          & 2594          & 2619          & 3797          & 5780          \\
% \textbf{ACL}    &                &               &               &               &               &               \\
% \textbf{EMNLP}  &                &               &               &               &               &               \\
% \textbf{}       &                &               &               &               &               &               \\ \bottomrule
% \end{tabular}}
% \end{table}

\subsection{Data Types}

% We develope a dataset, \texttt{ScholarLens}, including both human from two distinct perspectives: ``refined'' and ``synthesized''. The ``refined'' perspective focuses on enhancing and polishing existing drafts, while the ``synthesized'' perspective involves producing content based on provided texts.  In this section, we describe the process of curating \texttt{ScholarLens}, including considerations of academic aspects and the collection of LLM-generated text details.

We formalize a research paper as $\mathrm{P}=\left\{ \mathrm{T},\mathrm{A},\mathrm{C},\mathrm{R},\mathrm{MR} \right\}$ where $\mathrm{T}$, $\mathrm{A}$, and $\mathrm{C}$ represent the title, abstract, and main content, respectively, and $\mathrm{R}=\left\{ r_i \right\}$ denotes individual reviews, with $\mathrm{MR}$ representing the meta-review summarizing feedback from multiple reviewers.
Since the process of creating a research paper involves both author drafting and peer review stages, our dataset includes content from both the author and (meta-)reviewer roles. 
When creating LLM-generated text, we consider two perspectives: `refined', which enhances existing drafts, and `synthesized', which summarizes and generates content from provided texts.

% \paragraph{Author Role}
For the author role, we focus on abstract writing, as it is a key element for summarizing the paper and is easily accessible, making it ideal for this study. Specifically, we adopt the `refined' approach, where the original human-written abstract is input into the LLM to generate a refined version.
For the (meta-)reviewer roles, we focus on their comment content. To simulate human-written reviews and meta-reviews, the LLM-generated version primarily adopts the `synthesized' perspective. 
% Specifically, we provide a generation template as a guideline for the LLM.
The review process requires the full text of the paper as input, while the meta-review process includes all associated reviews of the paper.
All prompts used to create LLM-generated content are provided in the Appendix~\ref{app: prompts}.

% For author roles, our primary focus is on the integration of large language models (LLMs) in refining abstract content. The abstract serves as the core element of a paper, offering a concise summary that is both easily accessible and reliably extracted from public sources. Specifically, we employ an LLM as a `polisher', where the original human-written abstract is input into the model, which then generates a refined version. To ensure diversity and robustness in the refinement process, we design a prompt set, consisting of five distinct prompts. One prompt is randomly selected during each refinement cycle. 

% \paragraph{(Meta-)Reviewer Roles}

% A technical reviewer provides detailed feedback on the paper, evaluating aspects such as research methods, writing structure, experimental results, and novelty. In contrast, a meta-reviewer synthesizes the comments from multiple reviewers, offering a comprehensive summary that directly influences the final recommendation regarding the paper's acceptance.
% To simulate the process of human-written reviews and meta-reviews, the LLM-generated version primarily adopts the `synthesized' perspective. Specifically, we provide a generation template format as a guideline for the LLM. The review process requires the full text of the paper as input, while the meta-review process necessitates the inclusion of all the reviews of the paper. Detials of the generation tempalte are also shown in the Appendix~\ref{app: prompts}.

\subsection{Data Collection Setup}

Considering the challenges associated with parsing full-text papers, typically in PDF format, and the high computational cost of generating LLM-based reviews from lengthy input data, the \texttt{ScholarLens} collection integrates pre-existing review data with self-constructed abstracts and meta-reviews.

For the self-constructed data, we first collect raw data from all main conference papers in ICLR up to 2019, totaling 2,831 papers, through the OpenReview website. 
This selection is motivated by two factors: first, ICLR's peer review process provides comprehensive and detailed (meta-)review data; second, by focusing on papers before 2019, we can assume that the source data remains entirely human-written, as it predates the release of ChatGPT. For each paper, we generate two types of LLM-generated content: LLM-refined abstracts and LLM-synthesized meta-reviews. Both types are created using three advanced closed-source LLMs: GPT-4o, Gemini-1.5~\cite{team2024gemini}, and Claude-3 Opus~\cite{claude3modelcard}. This ensures that for every human-written version of the content, there is a corresponding LLM-generated version from each of the three models.
For the pre-existing review data, we directly leverage the review data from ReviewCritique~\cite{du-etal-2024-llms}\footnote{Note that ReviewCritique includes LLM-generated reviews for only 20 papers.}, which incorporates the same three LLMs.
 Details of \texttt{ScholarLens} are in Appendix~\ref{app:dataset}, including statistics and LLM settings.

\section{\texttt{LLMetrica} Framework}
\label{sec: LLMetria}
In this section, we introduce the \texttt{LLMetrica} framework, designed to evaluate the penetration rate of LLM-generated content in scholarly writing and peer review. 
The framework includes rule-based metrics for assessing linguistic features and semantic similarity, as well as model-based detectors fine-tuned specifically to identify LLM-generated content within the scholarly domain.

% which includes rule-based metrics for evaluating linguistic features and semantic similarity, along with model-based detectors fine-tuned specifically to identify LLM-generated content within the scholarly domain.

% \subsection{Linguistic and Semantic Metrics}  
\subsection{Rule-Based Metrics: Preference}  
\label{sec:metric}
Rule-Based Metrics define a metric function $m$ to measure the feature value $v$ of an input text $x$, i.e., $v=m(x)$, enabling the comparison and evaluation of feature preferences in LLM-generated text. 
Specifically, we use 10 general linguistic feature metrics and design 4 specialized semantic feature metrics to capture both linguistic and semantic characteristics.

% Linguistic and semantic metrics are proposed to analyze and compare the structural, syntactic, and meaning-based features of texts, helping to identify LLM preferences and infer trends in its generated text.

\subsubsection{General Linguistic Features}

General linguistic features are applicable to all types of text and can be categorized into word-level, sentence-level, and other related metrics.
Specifically, word-level metrics include Average Word Length (AWL), Long Word Ratio (LWR), Stopword Ratio (SWR), and Type Token Ratio (TTR). Given the nature of scholarly writing, the threshold for `long word' is set at 10. 
For sentence-level metrics, we include Average Sentence Length (ASL), Dependency Relation Variety (DRV), and Subordinate Clause Density (SCD).
DRV quantifies the diversity of dependency relations within the text using Shannon entropy~\cite{lin1991divergence}, while SCD focuses on dependency relations such as `advcl', `ccomp', `xcomp', `relcl', and `acl'~\cite{nivre-etal-2017-universal}.
In addition, we incorporate Flesch Reading Ease (FRE)~\cite{farr1951simplification} to evaluate the overall readability of the text, Sentiment Polarity Score (PS, range: [-1, 1], negative→positive) to assess the sentiment, and Sentiment Subjectivity Score (SS, range: [0, 1], objective → subjective) to measure the degree of subjectivity or objectivity in the text. The implementation details of these metrics are provided in the Appendix~\ref{app:general}.

\subsubsection{Specific Semantic Features}
% applicable to all types of text and can be categorized into word-level, sentence-level, and other related metrics.
Inspired by \citet{du-etal-2024-llms}, which shows that human-written reviews have greater diversity and segment-level specificity than LLM-generated ones, we design four semantic metrics to analyze meta-reviews and reviews, focusing on overall semantic similarity and sentence-level specificity.

% \citet{du-etal-2024-llms} demonstrate that human-written reviews typically exhibit greater diversity and higher segment-level specificity compared to those generated by LLMs. Building on this insight, we conduct a quantitative analysis of MR and R, focusing specifically on two aspects: (i) the overall semantic similarity to the reference text, and (ii) sentence-level specificity at a finer granularity.

\paragraph{Overall Semantic Similarity}

We propose two semantic similarity metrics:
(i) \textbf{MRSim}: measures the similarity between the MR and its reference set R, defined as the average semantic similarity between MR and each review $r_i \in R$.
(ii) \textbf{RSim}: measures the similarity among reviews within R, defined as the maximum similarity among all pairs of reviews in R. The formulas are:
% The mathematical expressions are as follows:
\begin{equation}
\small{\mathrm{MRSim}=\frac{1}{\left| \mathrm{R} \right|}\sum_{r_i\in \mathrm{R}}{\mathrm{sim} \left( \mathrm{MR}, r_i \right)} }
\end{equation}
\begin{equation}
\small{\mathrm{RSim}=\underset{r_i,r_j\in \mathrm{R},r_i\ne r_j}{\max}\mathrm{sim} \left( r_i, r_j \right) }
\end{equation}
Using maximum similarity for RSim accounts for the fact that not all reviews in R are LLM-generated, as averaging could obscure key differences. Focusing on the maximum similarity highlights the strongest alignment, offering a more accurate measure of overall similarity.

% The use of maximum similarity for RSim considers that not all reviews in R are LLM-generated, so averaging could obscure key differences, while focusing on the maximum similarity captures the strongest alignment, providing a more accurate measure of overall similarity.

\begin{figure*}[ht]
    \centering
    % \vspace{1cm}
    \begin{minipage}{1\linewidth}
        \centering
        \includegraphics[width=1.0\linewidth]{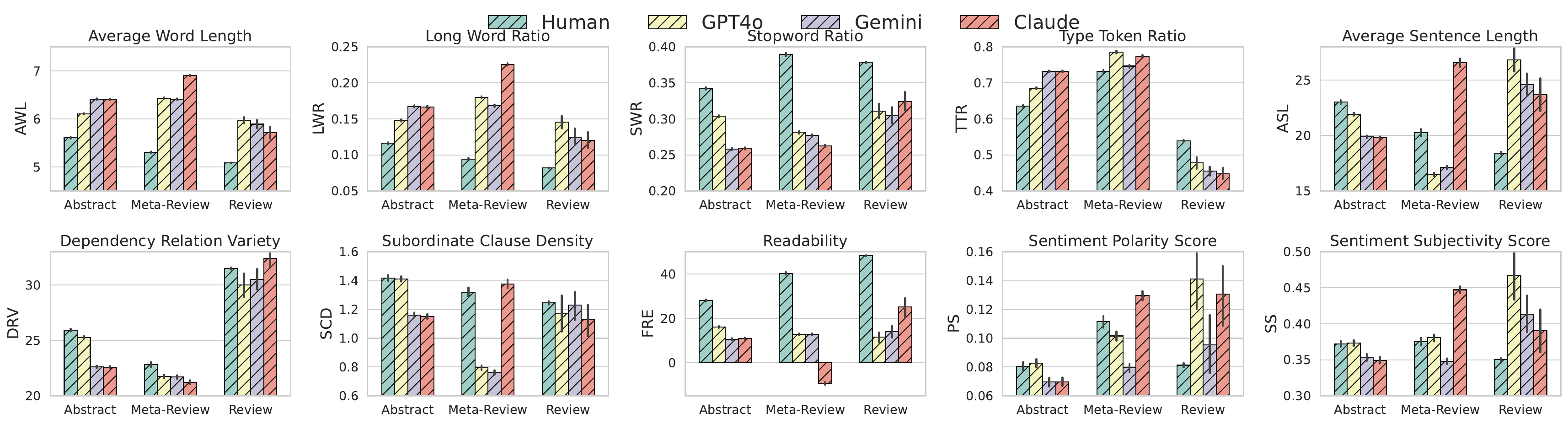}

        \subcaption{Comparison of Features for ALL Data Types}
    \end{minipage}

    \begin{minipage}{1\linewidth}
        \centering
        \scalebox{0.9}{
        \begin{tabular}{@{}l|lll@{}}
        \toprule
          Data Type  & ↑                        & ↓                          & →                 \\ \midrule
        \textbf{Abstract}    & 3 (\textbf{AWL}, \textbf{LWR}. TTR)         & 5 (\textbf{SWR}, ASL, DRV, SCD, \textbf{FRE}) & 2 (PS,SS)           \\
        \textbf{Meta-Review} & 3 (\textbf{AWL}, \textbf{LWR}. TTR)         & 3 (\textbf{SWR}, DRV, \textbf{FRE})           & 4 (ASL, SCD, PS, SS) \\
        \textbf{Review}      & 5 (\textbf{AWL}, \textbf{LWR}, ASL, PS, SS) & 3 (\textbf{SWR}, TTR, \textbf{FRE})           & 2 (DRV, SCD)         \\ \bottomrule
        \end{tabular}}

        \subcaption{Feature preference of LLM-generated text: ↑ indicates an increase across all LLMs, ↓ indicates a decrease, → indicates inconsistency. \textbf{Bold} denotes consistent trends across all data types.}
    \end{minipage}

    \caption{Comparison of Human-Written and LLM-Generated Text Based on \textbf{General Features} in \texttt{ScholarLens}}
    \label{fig:compare_general}
\end{figure*}

% \begin{figure*}[ht]
%     \centering
%     \includegraphics[width=1.0\linewidth]{fig/feature.pdf}
%     \caption{Caption}
%     \label{fig:enter-label}
% \end{figure*}

\paragraph{Sentence-Level Specificity}
Building on ITF-IDF~\cite{du-etal-2024-llms}\footnote{Unlike ITF-IDF~\cite{du-etal-2024-llms}, we only measure the SF-IRF within a single paper, considering the meta-review and review levels.} and the classic TF-IDF framework, we introduce the \textbf{SF-IRF} (\textbf{S}entence \textbf{F}requency-\textbf{I}nverse \textbf{R}everence \textbf{F}requency) metric to quantify the significance of sentences within a (meta-)review. 
Specifically, for a given target (meta)-review $r$ consisting of $n$ sentences, $\mathrm{SF}\text{-}\mathrm{IRF}\left( s,r,\mathrm{R}_{\mathrm{ref}} \right)$ captures the importance of a sentence $s$ in $r$ by considering: (i) its frequency of occurrence within $r$ (SF), and (ii) its rarity across the reference reviews $\mathrm{R}_\mathrm{ref}$ (IRF). The metric is formally defined as:
\begin{equation}
\footnotesize{
    \begin{aligned}
        \mathrm{SF}\text{-}\mathrm{IRF}\left( s,r,\mathrm{R}_{\mathrm{ref}} \right) &= \mathrm{SF}\left( s,r \right) \cdot \mathrm{IRF}\left( s, \mathrm{R}_{\mathrm{ref}} \right) \\
        &= \frac{O_{s}^{r}}{n} \cdot \log \left( \frac{m}{Q_{s}^{\mathrm{R}_{\mathrm{ref}}}} \right)
    \end{aligned}}
\end{equation}
Here, if $r$ represents a review, then $\mathrm{R}_{\mathrm{ref}}=\mathrm{R}-r$; if $r$ is a meta-review, then $\mathrm{R}_{\mathrm{ref}}=\mathrm{R}$. $O_{s}^{r}$ quantifies the ``soft'' occurrence of sentence $s$ within the target review $r$, while $Q_{s}^{\mathrm{R}_{\mathrm{ref}}}$ represents the ``soft'' count of reviews in $\mathrm{R}_{\mathrm{ref}}$ that contain the sentence $s$. Additionally, $m$ denotes the total number of reviews in $\mathrm{R}_{\mathrm{ref}}$. $O_{s}^{r}$ and $Q_{s}^{\mathrm{R}_{\mathrm{ref}}}$ are computed as follows:
\begin{equation}
\footnotesize{
    O_{s}^{r} = \sum\limits_{\tilde{s} \in r} \mathbb{I} \left( \mathop {\mathrm{sim}} \left( s, \tilde{s} \right) \geqslant t \right) \cdot \mathop {\mathrm{sim}}  \left( s, \tilde{s} \right)}
\end{equation}
\begin{equation}
\footnotesize{
    Q_{s}^{R_{\mathrm{ref}}} = \sum\limits_{\tilde{r} \in  R_{\mathrm{ref}}} \mathbb{I} \left( \mathop {\max \mathrm{sim}} \limits_{\tilde{s} \in \tilde{r}} \left( s, \tilde{s} \right) \geqslant t \right) \cdot \mathop {\max \mathrm{sim}} \limits_{\tilde{s} \in \tilde{r}} \left( s, \tilde{s} \right)}
\end{equation}
A segment $s$ is counted when its similarity exceeds the threshold $t$, with the corresponding similarity score.
% Following \citet{du-etal-2024-llms}, 
We use SentenceBERT~\cite{reimers-gurevych-2019-sentence} to calculate the all similarities, with $t$ set to 0.5.

% \subsection{Detection Models}
\subsection{Model-Based Detectors: Distinction}
\label{sec:detectors}
Model-based detectors are designed to train scholar-specific detection models 
$f$, capable of accurately identifying whether a scholarly input text $x$ is human-written or LLM-generated. We use our curated \texttt{ScholarLens} dataset to train these models, collectively referred to as ScholarDetect.

Specifically, we split abstracts and meta-reviews data within \texttt{ScholarLens} into training and test sets in a 7:3 ratio, based on the human-written versions. 
The corresponding LLM-generated content is partitioned accordingly, ensuring that each piece of LLM-generated text is paired with its human-written counterpart. 
All reviews data in \texttt{ScholarLens} are incorporated into the test set, ensuring a comprehensive evaluation.\footnote{Data statistics after splitting for model training and evaluation are in Table~\ref{tab:data_split}.}
All test sets
% , which include LLM-generated text from multiple LLM resources, 
serve as benchmarks to assess the performance of both baseline models and the trained ScholarDetect models.
We create three types of detection models based on the training data: one using only abstracts, one using only meta-reviews, and one using a hybrid of both.
To maintain class balance in the training data, we ensure a 1:1 ratio between human-written and LLM-generated version. We employ two strategies: one using a single LLM (GPT-4o, Gemini, or Claude), and another using a mixed-LLM approach, where each human-written piece is paired with LLM-generated content from a randomly selected model.
As a result, the ScholarDetect framework involves a total of 12 distinct detection models.

\section{Experiments}
\label{sec:Experiments}
We first demonstrate the effectiveness of rule-based metrics (\S\ref{sec: Features Comparison}) and ScholarDetect models (\S\ref{sec:ScholarDetect Evaluation}) in \texttt{LLMetrica}, then apply these methods to real-world conference data to assess and predict LLM penetration trends (\S\ref{sec:Temporal Analysis}). Finally, case studies are used to explore the specific differences between human-written and LLM-generated content (\S\ref{sec: Case Study}).

% case studies provide deeper insights into the distinguishing features of human-written vs. LLM-generated content (\S\ref{sec: Case Study}).

% we conduct case studies to gain deeper insights into the distinctive characteristics of human-written and LLM-generated content (\S\ref{sec: Case Study}).

\subsection{Features Comparison: Human vs LLM}
\label{sec: Features Comparison}
We apply the proposed rule-based metrics (\S\ref{sec:metric}) to  \texttt{ScholarLens} to compare the features of human-written and LLM-generated texts, and find that the feature preferences of LLM-generated texts can be effectively compared and evaluated.

% Results are shown in Figure~\ref{fig:compare_general} (general features) and Figure~\ref{fig:compare_specific} (specific features).

% \usepackage{multirow}
% \begin{table*}[]
% \begin{tabular}{c|c|ccc|ccc|ccc}
% \hline
% \multirow{2}{*}{}         & \multirow{2}{*}{Model} & \multicolumn{3}{c|}{Abstract} & \multicolumn{3}{c|}{Meta-Reivew} & \multicolumn{3}{c}{Reiview}                  \\ \cline{3-11} 
%                           &                        & Human   & LLM     & Overall   & Human    & LLM      & Overall    & Human & LLM   & \multicolumn{1}{c|}{Overall} \\ \cline{2-11} 
% \multirow{3}{*}{Baseline} & MAGE                   & 64.46   & 37.07   & 54.58     & 65.80    & 28.34    & 53.70      & 92.98 & 57.60 & \multicolumn{1}{c|}{87.95}   \\
%                           & RaiDetector            & 62.04   & 71.21   & 67.25     & 57.39    & 70.25    & 64.96      & 24.48 & 26.84 & \multicolumn{1}{c|}{25.68}   \\
%                           & HNDCDetector           & 54.59   & 85.85   & 78.43     & 62.63    & 76.74    & 71.33      & 86.39 & 54.90 & \multicolumn{1}{c|}{79.09}   \\ \hline
% \end{tabular}
% \caption{the F1 score performance of our fine-tuning model compare with two of current detect model}
% \label{tab:fine-detection}
% \end{table*}

% \usepackage{booktabs}
% \usepackage{multirow}
\begin{table*}[t]
\centering
\scalebox{0.64}{
\begin{tabular}{@{}l|p{1.3cm}|rrr|rrr|rrr|r@{}}
\toprule
\multirow{2}{*}{\textbf{Model}}                                  & \multirow{2}{*}{\parbox{1.3cm}{\textbf{LLM Source}}} & \multicolumn{3}{c|}{\textbf{Abstract}}                          & \multicolumn{3}{c|}{\textbf{Meta-Review}}                       & \multicolumn{3}{c|}{\textbf{Review}}                            & \multicolumn{1}{l}{\multirow{2}{*}{\textbf{Avg.}}} \\ \cmidrule(lr){3-11}
                                                                 &   \multirow{1}{*}                     & Human               & LLM                 & Overall             & Human               & LLM                 & Overall             & Human               & LLM                 & Overall             & \multicolumn{1}{l}{}                                  \\ \midrule
\textbf{MAGE}                                                    & \multirow{3}{*}{-}                 & 40.62               & 35.14               & 38.00               & 40.58               & 33.43               & 37.21               & 92.98               & 57.60                & 87.95               & 54.39                                                 \\
\textbf{RAIDetect}                                               &                                    & 49.51               & 78.36               & 69.71               & 38.42               & 73.50                & 62.94               & 24.48               & 26.85               & 25.68               & 52.78                                                 \\
\textbf{HNDCDetect}                                              &                                    & 54.59               & 85.85               & 78.43               & 57.93               & 83.88               & 76.69               & 86.39               & 54.90                & 79.09               & 78.07                                                 \\ \midrule
\multirow{4}{*}{\textbf{$\text{ScholarDetect}_{\text{Abs}}$}}    & GPT-4o                             & 97.02\small ±0.50          & 99.02\small ±0.15          & 98.53\small ±0.23          & 87.65\small ±3.01          & 95.06\small ±1.48          & 92.95\small ±2.00          & 97.44\small ±0.20          & 79.97\small ±1.96          & 95.45\small ±0.37          & 95.64                                                 \\
                                                                 & Gemini                             & 84.40\small ±3.11          & 93.48\small ±1.59          & 90.80\small ±2.13          & 90.62\small ±1.98          & 95.06\small ±1.21          & 92.93\small ±1.63          & 96.41\small ±0.64          & 68.68\small ±7.26          & 93.56\small ±1.19          & 92.43                                                 \\
                                                                 & Claude                             & 91.61\small ±1.09          & 96.90\small ±0.46          & 95.47\small ±0.65          & 83.79\small ±4.74          & 93.02\small ±2.60          & 90.25\small ±3.40          & 95.69\small ±0.89          & 58.84\small ±12.55         & 92.20\small ±1.68          & 91.97                                                 \\
                                                                 & Mix                                & \underline{97.94\small ±0.16}          & \underline{99.32\small ±0.05}          & \underline{98.98\small ±0.07}          & 93.84\small ±0.23          & 97.81\small ±0.09          & 96.76\small ±0.13          & 97.56\small ±0.20          & 81.16\small ±1.92          & 95.68\small ±0.37          & 97.14                                                 \\ \midrule
\multirow{4}{*}{\textbf{$\text{ScholarDetect}_{\text{Meta}}$}}   & GPT-4o                             & 78.59\small ±1.64          & 90.47\small ±1.19          & 86.81\small ±1.45          & 99.53\small ±0.13          & 99.84\small ±0.04          & 99.76\small ±0.06          & 97.98\small ±0.21          & 84.95\small ±1.89          & 96.44\small ±0.39          & 94.34                                                 \\
                                                                 & Gemini                             & 80.13\small ±2.71          & 91.45\small ±1.56          & 88.05\small ±2.02          & 97.70\small ±1.86          & 99.19\small ±0.67          & 98.80\small ±0.98          & 97.65\small ±0.30          & 81.90\small ±2.69          & 95.83\small ±0.54          & 94.23                                                 \\
                                                                 & Claude                             & 62.42\small ±13.17         & 69.60\small ±16.38         & 66.93\small ±15.79         & 86.72\small ±7.94          & 94.15\small ±3.67          & 91.88\small ±5.02          & 97.82\small ±0.62          & 83.23\small ±5.56          & 96.14\small ±1.11          & 84.98                                                 \\
                                                                 & Mix                                & 84.20\small ±3.59          & 94.12\small ±2.24          & 91.47\small ±2.89          & \underline{99.84\small ±0.10}          & \underline{99.95\small ±0.03}          & \underline{99.92\small ±0.05}          & \textbf{99.52\small ±0.16} & \textbf{96.86\small ±1.08} & \textbf{99.17\small ±0.28} & 96.85                                                 \\ \midrule
\multirow{4}{*}{\textbf{$\text{ScholarDetect}_{\text{Hybrid}}$}} & GPT-4o                             & 97.69\small ±0.35          & 99.23\small ±0.13          & 98.84\small ±0.19          & 99.33\small ±0.17          & 99.78\small ±0.06          & 99.67\small ±0.08          & 97.73\small ±0.16          & 82.72\small ±1.42          & 95.98\small ±0.28          & \underline{98.16}                                                 \\
                                                                 & Gemini                             & 85.88\small ±0.19          & 94.32\small ±0.10          & 91.90\small ±0.13          & 97.84\small ±1.19          & 99.25\small ±0.43          & 98.89\small ±0.63          & 97.69\small ±0.31          & 82.29\small ±2.83          & 95.91\small ±0.56          & 95.56                                                 \\
                                                                 & Claude                             & 85.61\small ±1.67          & 94.11\small ±0.84          & 91.64\small ±1.13          & 96.49\small ±1.22          & 98.77\small ±0.44          & 98.18\small ±0.65          & 97.48\small ±0.26          & 80.35\small ±2.44          & 95.53\small ±0.47          & 95.13                                                 \\
                                                                 & Mix                                & \textbf{98.11\small ±0.35} & \textbf{99.37\small ±0.11} & \textbf{99.06\small ±0.17} & \textbf{99.88\small ±0.00} & \textbf{99.96\small ±0.00} & \textbf{99.94\small ±0.00} & \underline{98.06\small ±0.18}          & \underline{85.69\small ±1.53}          & \underline{96.59\small ±0.32}          & \textbf{98.53}                                        \\ \bottomrule
\end{tabular}}
\caption{Detection performance comparison of baseline models and ScholarDetect. \textbf{Bold} denotes the best performance, and \underline{underlined} denotes the second-best. ``Avg.'' shows the average overall score across the three test data types.}
\label{tab:fine-detection}
\end{table*}

\paragraph{General Linguistic Features}

Figure~\ref{fig:compare_general} shows trends in the characteristics of LLM-generated texts, with slight variations across different data types. Each metric reflects the consistency of features across texts generated by the three LLMs in at least one data type, demonstrating the `comparability' effectiveness of the chosen metrics.
Moreover, regardless of the data type or LLM used, LLM-generated texts consistently show higher values for Average Word Length (AWL) and Long Word Ratio (LWR), and lower values for Stopword Ratio (SWR) and Readability (FRE). This suggests that LLM-generated texts tend to use longer words, avoid excessive stopwords, and have lower readability.
For shorter text types, such as abstracts and meta-reviews, the observed increase in Type Token Ratio (TTR) reflects greater lexical diversity in LLM-generated texts.
This may be due to the conciseness inherent in short-form LLM-generated content. In contrast, for longer reviews, TTR decreases, potentially highlighting the limitations of LLMs in producing long-form content~\cite{wang-etal-2024-m4, wu2025survey}. Longer reviews may lack specificity~\cite{du-etal-2024-llms}, resulting in redundancy and repetitive segments.
Additionally, LLM-generated reviews tend to be more positive and subjective, suggesting a more favorable tone and less neutral objectivity. This aligns with \citet{jin-etal-2024-agentreview}, who found that LLM-generated reviews generally assign higher scores and show a higher acceptance rate.

\begin{figure}[t]
    \centering
    \begin{minipage}{0.55\linewidth}
        \centering
        \includegraphics[width=\linewidth]{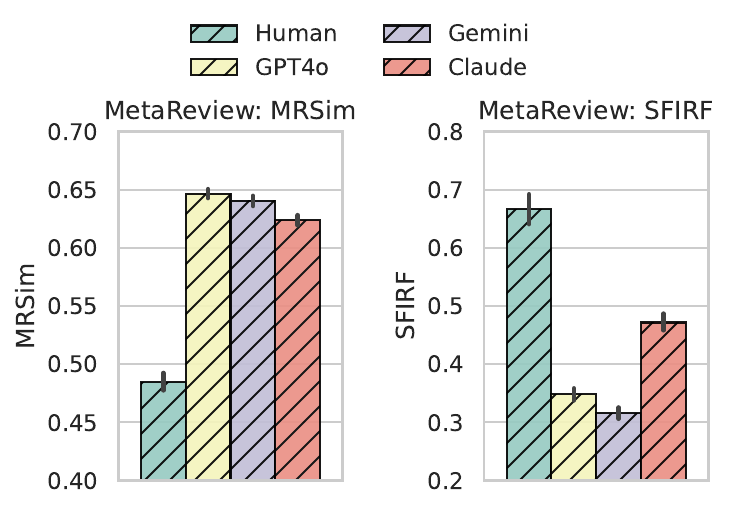}
        \captionsetup{font=footnotesize}
        \subcaption{Meta-Review}
    \end{minipage}
    \hfill
    \begin{minipage}{0.4\linewidth}
        \centering
        \includegraphics[width=\linewidth]{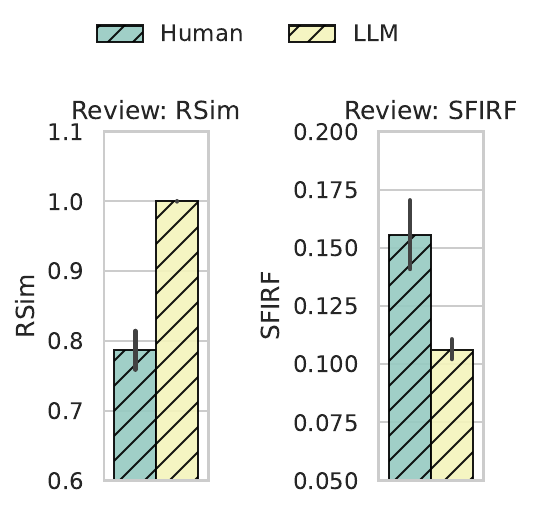}
        \captionsetup{font=footnotesize}
        \subcaption{Review}
    \end{minipage}
    \hfill
    \caption{Comparison of Human-Written and LLM-Generated Text Based on \textbf{Specific features} for Review and Meta-Review.}
    \label{fig:compare_specific}
\end{figure}

\paragraph{Specific Semantic Features}
Figure~\ref{fig:compare_specific} shows the preferences of human-written and LLM-generated texts in both meta-reviews and reviews, based on four specific semantic features.
Notably, Since each LLM generates only one review per paper in \texttt{ScholarLens}, while each paper usually has multiple reviews, we combine reviews from all three LLMs into a unified set, so comparisons do not distinguish between them.
Comparative results show that LLM-generated meta-reviews exhibit higher semantic similarity to the referenced reviews, with lower sentence specificity.
This suggests that sentences within LLM-generated meta-reviews are more semantically similar to each other (prone to redundancy) and tend to mirror the content of the referenced reviews. 
A similar trend is observed for reviews, where the two specific features also show consistent patterns.
It is important to note that for the reviews, we assume all are LLM-generated in this experiment, which may amplify the differences in these semantic features. In reality, having more than two LLM-generated reviews per paper may be uncommon, which would likely reduce the observed disparity.

% Figure~\ref{fig:compare_specific} shows that LLM-generated meta-reviews exhibit higher semantic similarity to the referenced reviews. Additionally, the specificity of segments is lower, indicating that segments within LLM-generated meta-reviews are more semantically similar to each other (prone to redundancy) and tend to mirror the content of the referenced reviews. Similarly, for reviews, the trends observed in the two specific features are also consistent.
% It is important to note that for the reviews, we assume all are LLM-generated in this experiment, which may amplify the differences in the two semantic features. In reality, having more than two LLM-generated reviews per paper is rare, which would likely reduce the observed disparity.

% AgentReview~\cite{jin-etal-2024-agentreview}.

\subsection{ScholarDetect Evaluation: Detectability}
\label{sec:ScholarDetect Evaluation}
We evaluate the trained model-based detectors, ScholarDetect (\S\ref{sec:detectors}), on the \texttt{ScholarLens} test sets and find that scholarly LLM-generated texts can be effectively identified.

\paragraph{Experimental Setup}
\textbf{(i)} \textbf{Training Setup}: 
We adopt Longformer~\cite{Beltagy2020Longformer} as the base model for training our ScholarDetect detection models, as it has shown competitive performance among pretrained language models~\cite{li-etal-2024-mage, cheng2024beyond}. Specifically, we train for five epochs in each configuration of the training set, using a learning rate of 2-e5.
\textbf{(ii)} \textbf{Metric}: 
For evaluation metrics, we report the F1 score for each class (human-written and LLM-generated), as well as the overall weighted F1 score to account for class imbalance. 
Each experimental setup (training data type and LLM-generated text source) is evaluated through three random trials, and we report the average performance along with the standard deviation. 
\textbf{(iii)} \textbf{Baselines}:
We compare the performance of three advanced detection model baselines: MAGE~\cite{li-etal-2024-mage}, RAIDetect~\cite{dugan-etal-2024-raid}, and HNDCDetect~\cite{cheng2024beyond}.
\textbf{(iv)} \textbf{Test Sets}:
All models are evaluated on test sets from three data types: abstract, meta-review, and review, with the first two being shorter texts and reviews being long-form. 
The LLM-generated data includes tasks such as refinement (for abstracts) and summarization (for meta-reviews and reviews).

\paragraph{Experimental Results}
The detection performance comparison results are presented in Table~\ref{tab:fine-detection}. Our trained ScholarDetect models consistently outperform the existing advanced baseline models, underscoring the importance of developing detection systems specifically tailored for the scholarly domain.
The training approach that combines mixed LLM sources and hybrid data types yields the best overall performance, demonstrating robustness across various LLM sources and data types.
Interestingly, the model trained on meta-reviews performs best when tests on reviews, likely because both data types share a similar comment-based focus and offer a ``synthesized'' perspective in LLM-generated text. 
This is further supported by ScholarDetect$_\text{Abs}$, which struggles to identify LLM-generated reviews when trained only on abstracts (F1: LLM < Human).
Additionally, when trained on a single LLM source, the GPT-4o-based detectors show the strongest generalization, especially on the abstract test set. 
Most ScholarDetect models outperform human-written text in detecting LLM-generated content on the meta-review and review test sets, but the reverse is true for the review test set.

% Here we present the results of training longformer. The loss rate of the trained model is shown in the table~\ref{tab:longformer_training_loss}, and the model is named according to the timestamp at which it was trained.

% We used our trained models to detect the metareview data from ICLR 2020-2024. We found that the AI generation rate was low from 2020 to 2023, but there was a sharp increase in 2024. Although the training was based on metareview data, we also applied our models to analyze abstracts, where a similar sharp rise was observed in 2024. Please see figure~\ref{fig:AI-generated metareview percentage} and figure~\ref{fig:AI-processed abstract percentage} for details.

\begin{figure}[t]
    \centering
    \begin{minipage}{0.98\linewidth}
        \centering
        \includegraphics[width=\linewidth]{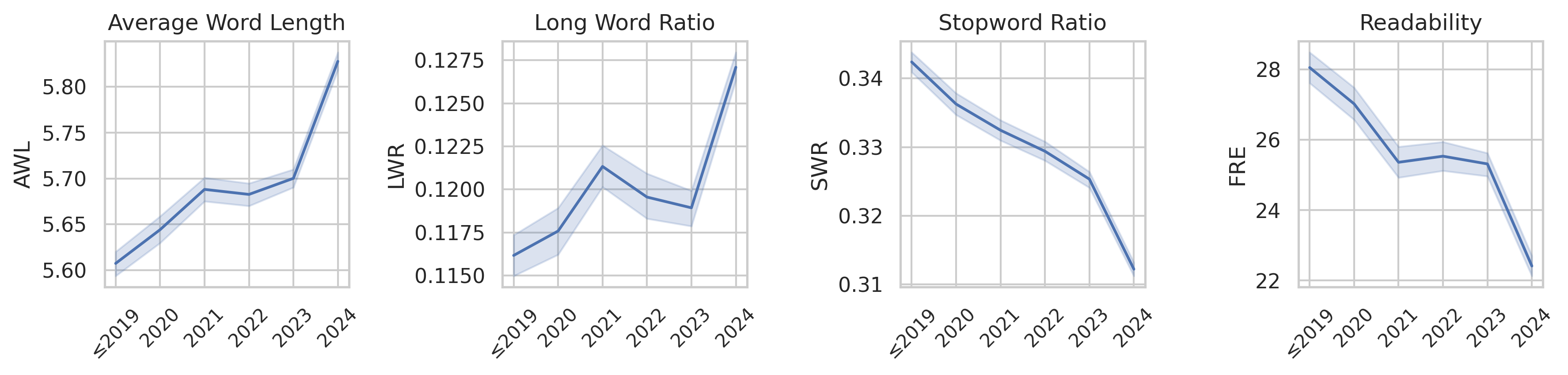}
        \captionsetup{font=footnotesize}
        \subcaption{Abstract}
    \end{minipage}
    \hfill
    \begin{minipage}{0.98\linewidth}
        \centering
        \includegraphics[width=\linewidth]{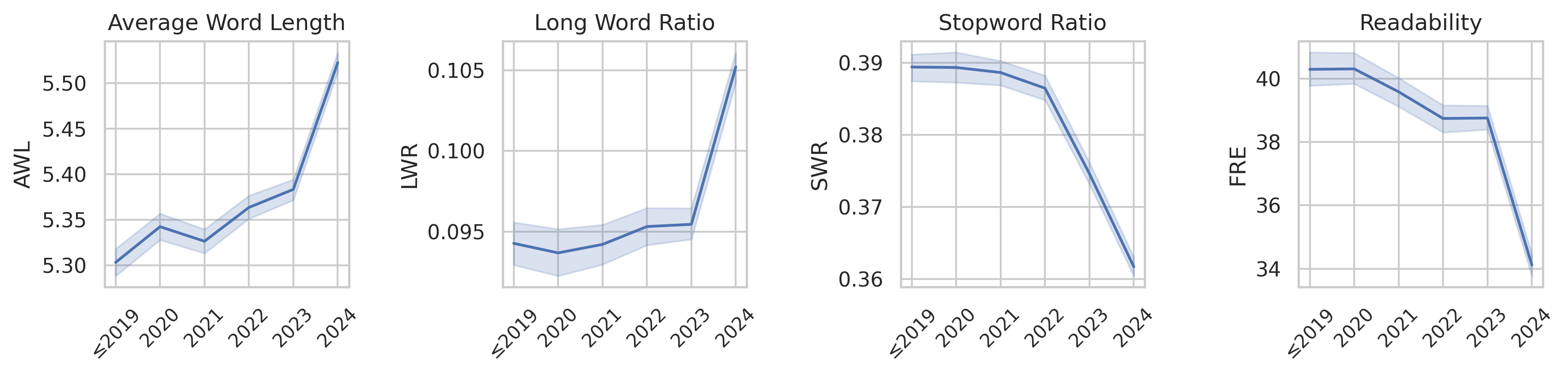}
        \captionsetup{font=footnotesize}
        \subcaption{Meta-Review}
    \end{minipage}
    \hfill
    \begin{minipage}{0.98\linewidth}
        \centering
        \includegraphics[width=\linewidth]{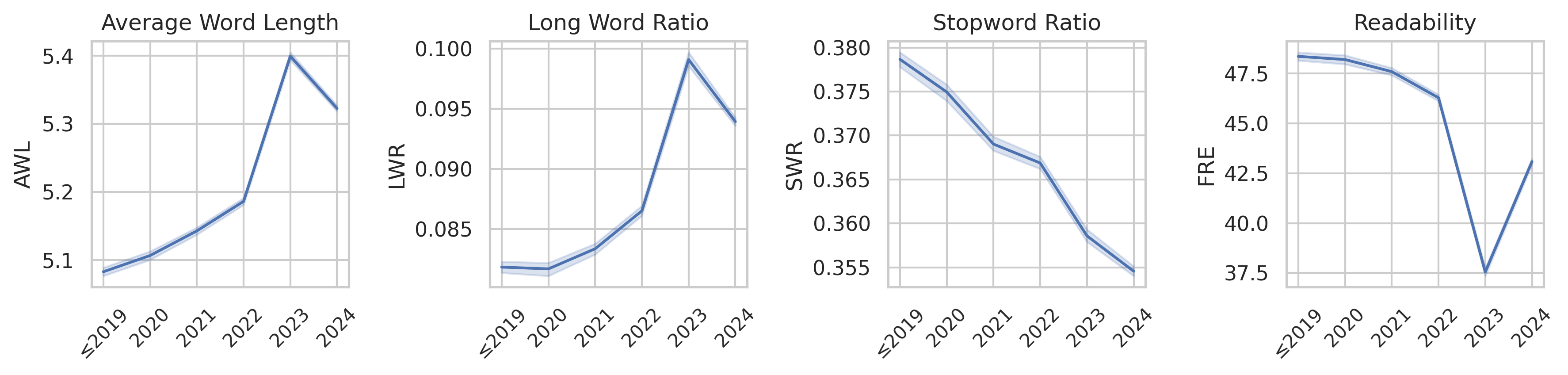}
        \captionsetup{font=footnotesize}
        \subcaption{Review}
    \end{minipage}
    \hfill

    \caption{Temporal trends based on \textbf{four robust general linguistic metrics}.
    % \red{$\boldsymbol{\uparrow \downarrow \rightarrow}$} represents the feature preference of LLM-generated text, as identified in the comparisons in Figures~\ref{fig:compare_general} and~\ref{fig:compare_specific}. \green{$\boldsymbol{\checkmark}$} indicates alignment between the feature trend and preference, signifying increased LLM penetration.
    }
    \label{fig:trend_general}
\end{figure}

% \begin{figure}[t]
%     \centering
%     \begin{minipage}{1\linewidth}
%         \centering
%         \includegraphics[width=\linewidth]{fig/specific_trend_meta.PNG}
%         \captionsetup{font=footnotesize}
%         \subcaption{Meta-Review}
%     \end{minipage}
%     \hfill
%     \begin{minipage}{1\linewidth}
%         \centering
%         \includegraphics[width=\linewidth]{fig/specific_trend_review.PNG}
%         \captionsetup{font=footnotesize}
%         \subcaption{Review}
%     \end{minipage}
%     \hfill
%     \caption{Temporal trends based on \textbf{specific semantic metrics}.}
%     \label{fig:trend_specific}
% \end{figure}

\begin{figure}[t]
    \centering
    \begin{minipage}{0.49\linewidth}
        \centering
        \includegraphics[width=\linewidth]{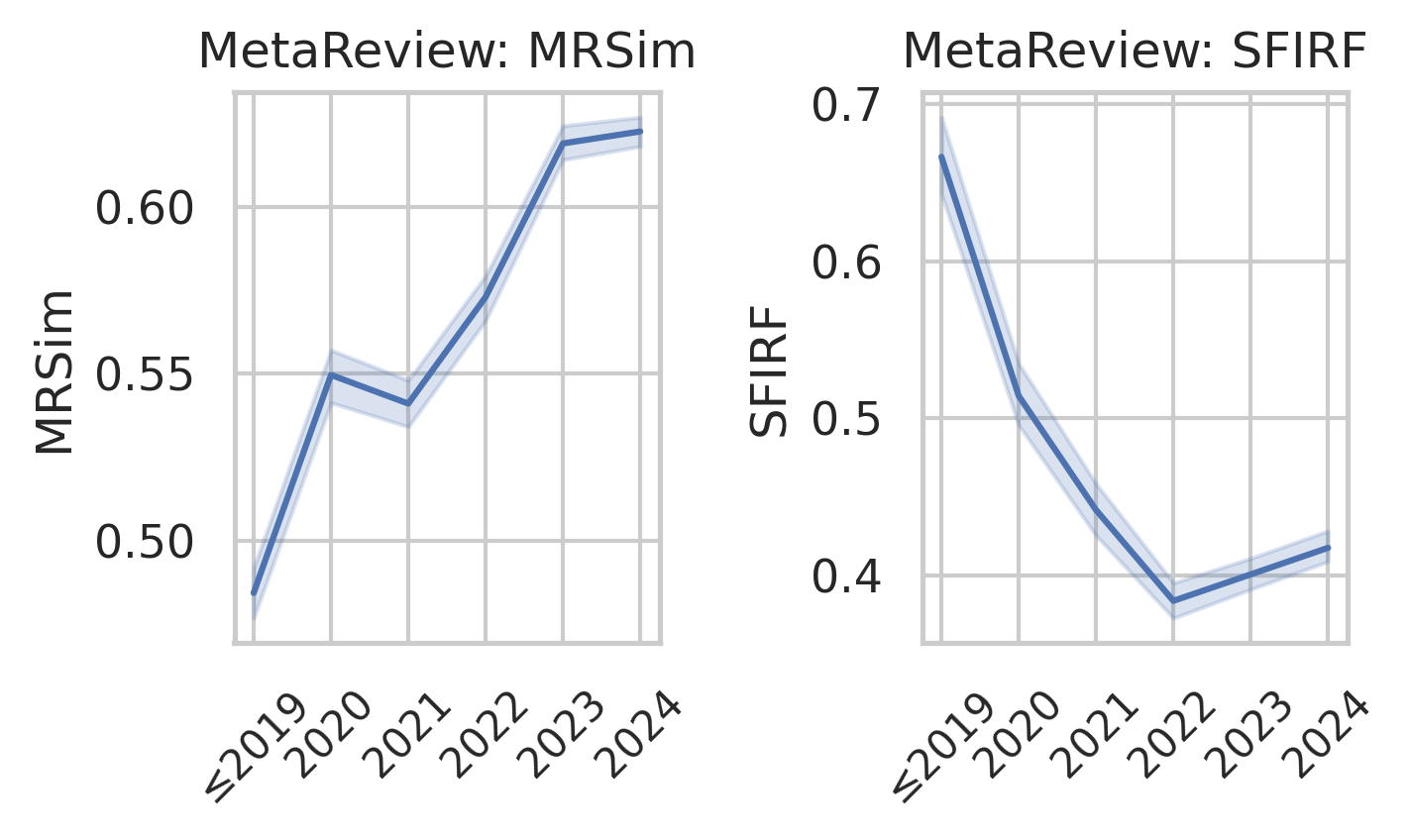}
        \captionsetup{font=footnotesize}
        \subcaption{Meta-Review}
    \end{minipage}
    \hfill
    \begin{minipage}{0.49\linewidth}
        \centering
        \includegraphics[width=\linewidth]{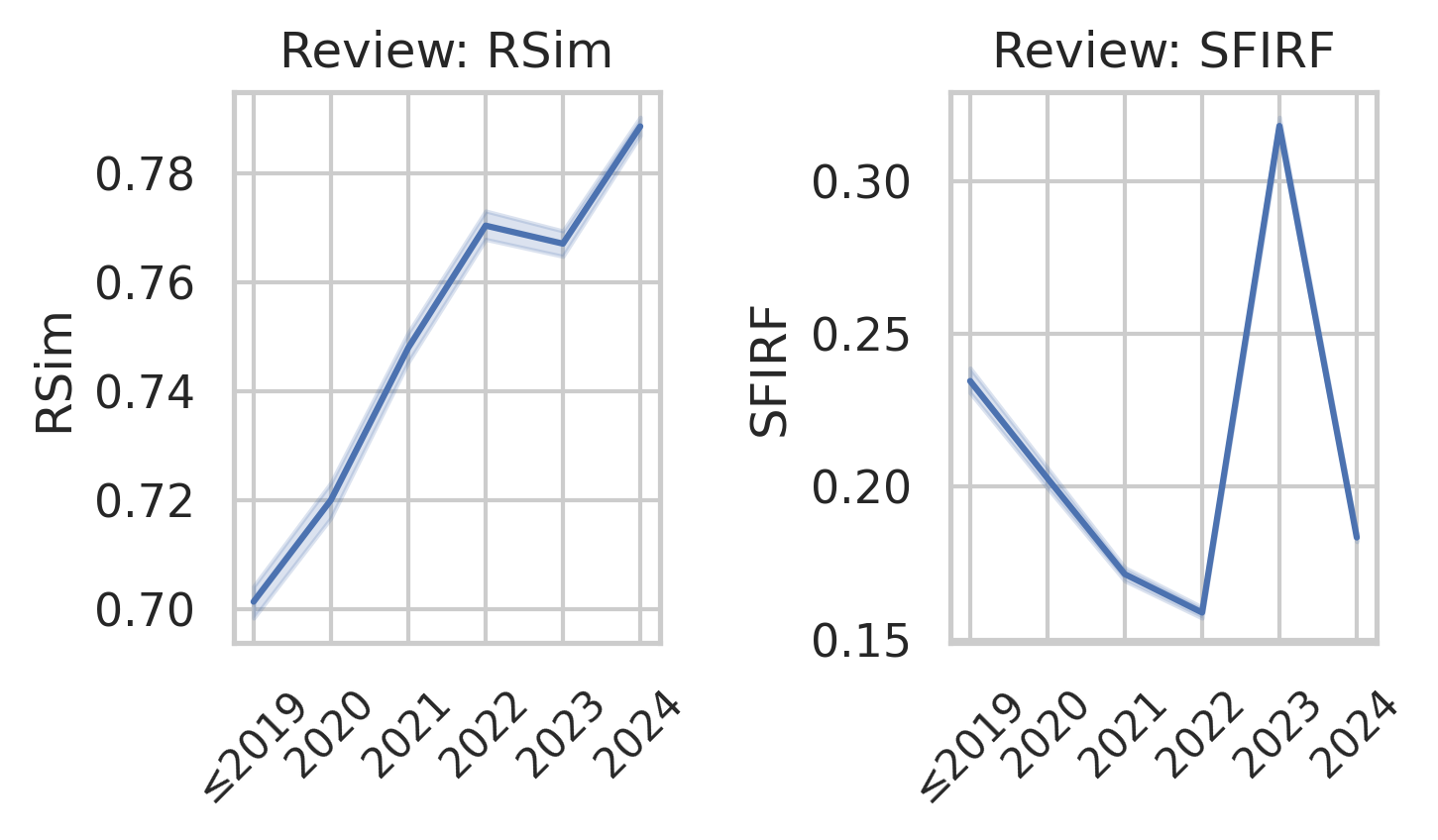}
        \captionsetup{font=footnotesize}
        \subcaption{Review}
    \end{minipage}
    \hfill
    \caption{Temporal trends based on \textbf{specific semantic metrics}.}
    \label{fig:trend_specific}
\end{figure}

\begin{figure}[t]
    \centering
    \includegraphics[width=1\linewidth]{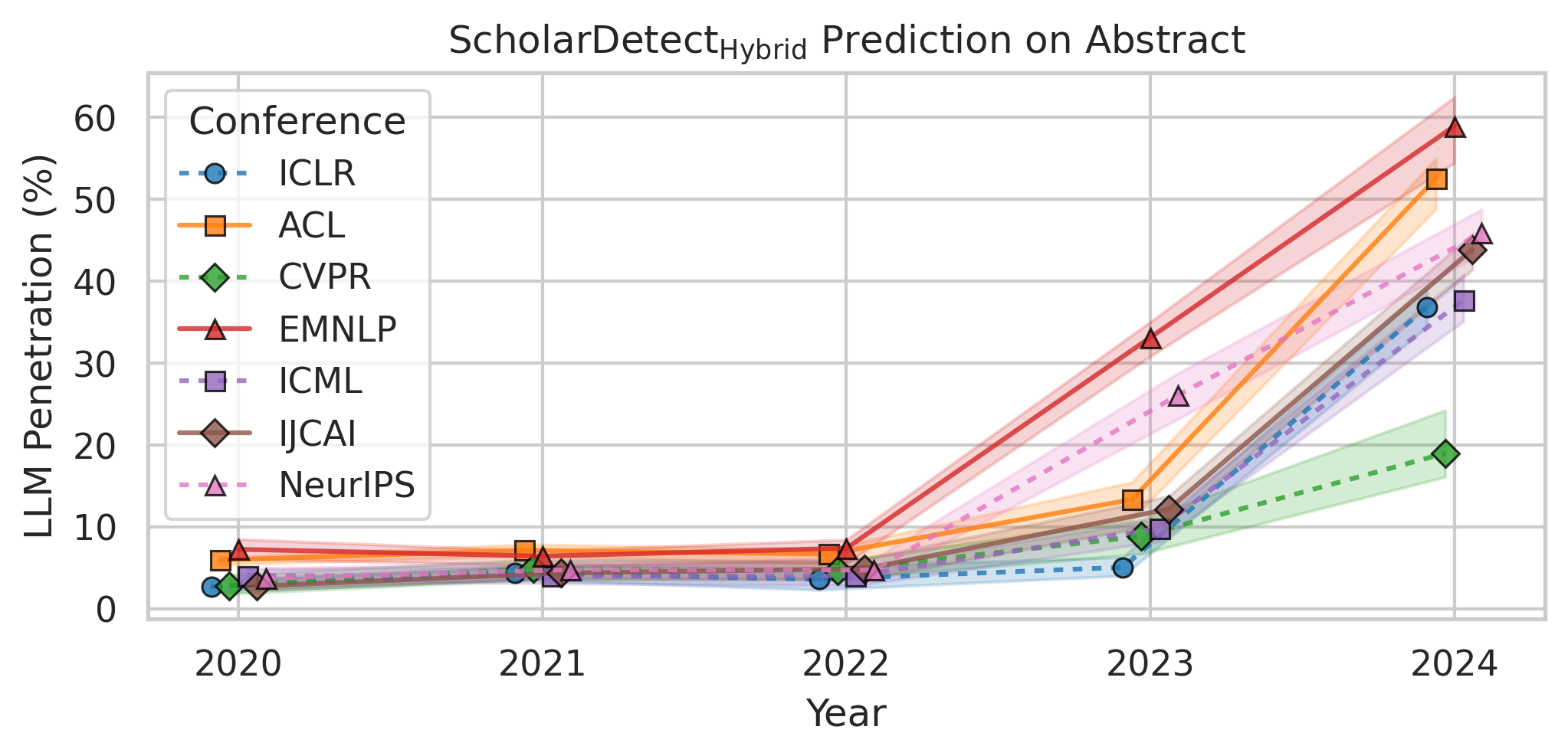}
    \caption{Abstarct: Trend based on detection model.}
    \label{fig:trend_detect_abs}
\end{figure}

\begin{figure}[t]
    \centering
    \includegraphics[width=1\linewidth]{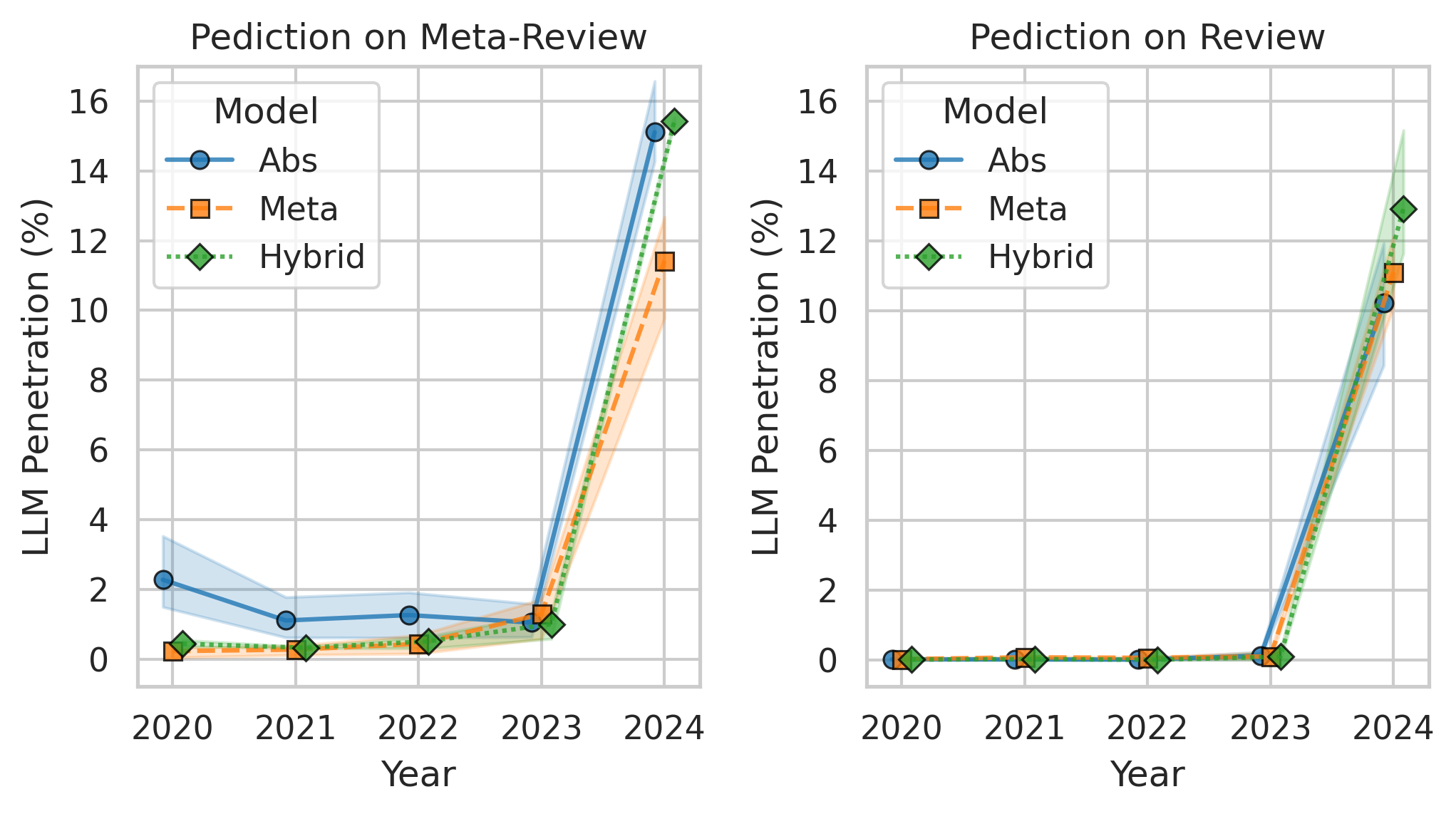}
    \caption{Abstarct: Trend based on detection model.}
    \label{fig:trend_detect_review}
\end{figure}

\subsection{LLM Penetration: Temporal Analysis}
\label{sec:Temporal Analysis}
We apply the proposed rule-based metrics and model-based detectors to assess and detect LLM penetration in recent scholarly texts (up to 2024), including abstracts, meta-reviews, and reviews. 

% We apply the proposed rule-based metrics and model-based detectors to assess and detect LLM penetration in recent scholarly texts (up to 2024), including abstracts, meta-reviews, and reviews.
% specifically, for rule-based metrics, we only adopt the four most robust general linguistic metrics across three data types.

\paragraph{Trend in Rule-based Evaluation}
For the general linguistic metrics, we use only the four most robust (AWL, LWR, SWR, FRE), which show consistent preferences across the three data types, and we adopt all four specific semantic metrics.
Figure~\ref{fig:trend_general} illustrates the trend in general linguistic features across three data types in ICLR, while Figure~\ref{fig:trend_specific} shows the trend in specific semantic features for meta-reviews and reviews. 
Almost all the metrics show consistent LLM preference trends across their associated data types, with an overall year-on-year increase, supporting the rising trend of LLM penetration in scholarly writing.
Interestingly, among these metrics used to evaluate reviews, four show anomalous trend changes in 2023, highlighting the difficulties of using rule-based metrics to track LLM penetration in the complex and varied nature of review data.

\paragraph{Trend in Model-based Detection}
Based on the performance shown in Table~\ref{tab:fine-detection} and the available evaluation data, we select ScholarDetect$_\text{Hybrid}$, which performs best on abstracts, to detect instances of LLM-assisted writing in seven conference abstracts.
% Based on the performance presented in Table~\ref{tab:fine-detection} and the available evaluation data sources, we adopt ScholarDetect$_\text{Hybrid}$, which performs best on abstracts, to probe seven conference abstracts for identifying instances of LLM-assisted writing.
Additionally, we utilize three variants of ScholarDetect (Abs, Meta, Hybrid) to analyze all ICLR meta-reviews and reviews. The detected LLM penetration rates (i.e., the proportion of text predicted to be LLM-generated) are presented in Figures~\ref{fig:trend_detect_abs} and~\ref{fig:trend_detect_review}.
In the abstract evaluation data, the LLM penetration rate across all involved conferences increases starting in 2023 and continues to rise in 2024, likely driven by ChatGPT’s initial release in November 2022 and its subsequent updates.
In contrast, a noticeable increase appears in 2024 for comment-based data, particularly in reviews, although the overall rate remains lower than in abstracts.
This may be attributed to the 2023 update of ChatGPT\footnote{\href{https://help.openai.com/en/articles/6825453-chatgpt-release-notes}{ChatGPT — Release Notes}}, which enabled PDF uploads and content analysis, as well as the higher standards required for LLM-generated content in reviews, which limit the penetration rate.
Specifically, ScholarDetect$_\text{Hybrid}$ predicts the highest LLM penetration rate for two comment-based data types in 2024.  For shorter meta-review texts, ScholarDetect$_\text{Abs}$'s rate is close to ScholarDetect$_\text{Hybrid}$ but higher than ScholarDetect$_\text{meta}$. We hypothesize this is due to the greater role of LLMs in refining these texts. Based on insights from ~\citet{cheng2024beyond}, we propose a fine-grained LLM-generated text detection approach using three-class role recognition (human-written, LLM-synthesized, LLM-refined) for meta-reviews. Our results show that the LLM-refined role plays a more dominant part in LLM penetration.\footnote{Experimental details of the three-class LLM role recognition are in Appendix~\ref{app:Fine-Grained}.}

\begin{table}[t]
    \centering
    \scalebox{0.8}{
    \begin{tabular}{@{}l|p{8cm}}
    \toprule
    \textbf{POS} & \textbf{ Top 10 GPT-4o Preferred Words in Meta-Reviews} \\ \midrule
    \textbf{\small NOUN}   & \small \underline{\textbf{refinement}}, \underline{\textbf{advancements}}, \underline{\textbf{methodologies}}, \underline{\textbf{articulation}}, 
    \textbf{highlights}, \underline{reliance}, \underline{\textbf{enhancement}}, \underline{\textbf{underpinnings}}, \underline{\textbf{enhancements}}, \underline{\textbf{transparency}}  \\ \midrule
    \textbf{\small VERB}   & \small \underline{enhance}, \underline{enhancing}, deemed, \underline{\textbf{showcasing}}, express, \underline{offering}, \underline{enhances}, \underline{\textbf{recognizing}}, commend, praised                  \\ \midrule
    \textbf{\small ADJ}  & \small \underline{\textbf{innovative}}, \underline{\textbf{collective}}, \underline{enhanced}, \underline{\textbf{established}}, \underline{notable}, \underline{outdated}, varied, \underline{undefined}, \underline{\textbf{comparative}}, \underline{\textbf{noteworthy}}
            \\ \midrule
    \textbf{\small ADV}  & \small \underline{\textbf{collectively}}, \underline{\textbf{inadequately}}, \underline{\textbf{reportedly}}, \underline{\textbf{comprehensively}}, \underline{robustly}, \underline{\textbf{occasionally}}, \underline{\textbf{predominantly}}, \underline{notably}, \underline{\textbf{innovatively}}, \underline{\textbf{effectively}}
          \\ 
  \bottomrule
  \toprule
    \textbf{POS} & \textbf{ Top 10 GPT-4o Preferred Words in Abstracts} \\ \midrule
    \textbf{\small NOUN}   & \small abstract, \underline{\textbf{advancements}}, realm, \underline{\textbf{alterations}}, aligns, \underline{\textbf{methodologies}}, \underline{clarity}, \underline{\textbf{adaptability}}, \underline{surpasses}, \underline{\textbf{examination}}
  \\ \midrule
    \textbf{\small VERB}   & \small \underline{enhancing}, \underline{\textbf{necessitates}}, \underline{\textbf{necessitating}}, \underline{featuring}, \underline{revised}, \underline{\textbf{influenced}}, \underline{\textbf{encompassing}}, \underline{enhances}, \underline{\textbf{showcasing}}, \underline{surpasses}
                  \\ \midrule
    \textbf{\small ADJ}  & \small \underline{\textbf{innovative}}, \underline{\textbf{exceptional}}, \underline{pertinent}, \underline{intricate}, \underline{pivotal}, \underline{\textbf{necessitate}}, \underline{\textbf{distinctive}}, \underline{enhanced}, akin, potent
            \\ \midrule
    \textbf{\small ADV}  & \small \underline{\textbf{inadequately}}, \underline{\textbf{predominantly}}, \underline{\textbf{meticulously}}, \underline{\textbf{strategically}}, \underline{notably}, abstract, swiftly, \underline{\textbf{additionally}}, \underline{adeptly}, \underline{thereby}
          \\ 
  \bottomrule
    \end{tabular}
    }
    \caption{Top-10 LLM-preferred words in GPT-4o-generated vs. human-written meta-reviews and abstracts. \textbf{Bold} denotes \textit{long words}, and \underline{underlined} denotes \textit{complex-syllabled words}.}
    \label{tab:word_case}
\end{table}

\subsection{Case Study}
\label{sec: Case Study}
To investigate the specific differences between LLM-generated and human-written text, we focus on GPT-4o, conducting case studies at both the word and pattern levels. 
(i) At the word level, we design a Two-Sample t-test based on word proportions~\cite{cressie1986use, 10.1093/biomet/34.1-2.28}\footnote{\url{https://www.statology.org/two-sample-t-test/}. Method details of case studies and additional results are in the Appendix~\ref{app: cs-details}.} to identify the LLM-preferred words.
Table~\ref{tab:word_case} shows the top 10 preferred words in four key part-of-speech (POS)
% \footnote{\lz{We have tried various POS tagging tools such as NLTK, spaCy, Stanford CoreNLP and others. All the tools that we have tried have the problem of mislabeling the POS of minor words.}} 
categories from GPT-4o-generated abstracts and meta-reviews, compared to those in human-written versions. We find that LLMs tend to generate \textit{long words} ($\geq$ 10 letters) and \textit{complex-syllabled words} ($\geq$ 3 syllables)~\cite{gunning1952technique}. This further supports the reliability of the four general linguistic metrics for assessing LLM penetration.
Moreover, GPT-4o shows a strong preference for the word `enhance' in scholarly writing and peer reviews, with its variants appearing in the top 10 list.
(ii) At the pattern level, manual inspection of paired data samples from comment-based data\footnote{Conducted by one of the authors on 100 paired meta-reviews and 20 paired reviews.}, followed by automated evaluation of the full dataset, reveals that human-written (meta-)reviews exhibit: 
\textit{personability}, frequently using the first person to express opinions; \textit{interactivity}, often incorporating questions; and \textit{attention to detail}, citing relevant literature to support arguments.

\section{Conclusion and Suggestions}

Our work, including the creation of \texttt{ScholarLens} and the proposal of \texttt{LLMetrica}, provides methods for assessing LLM penetration in scholarly writing and peer review. By incorporating diverse data types and a range of evaluation techniques, we consistently observe the growing influence of LLMs across various scholarly processes, raising concerns about the credibility of academic research. As LLMs become more integrated into scholarly workflows, it is crucial to establish strategies that ensure their responsible and ethical use, addressing both content creation and the peer review process. 

Despite existing guidelines restricting LLM-generated content in scholarly writing and peer review,\footnote{\href{https://aclrollingreview.org/acguidelines\#-task-3-checking-review-quality-and-chasing-missing-reviewers}{Area Chair} \&  \href{https://aclrollingreview.org/reviewerguidelines\#q-can-i-use-generative-ai}{Reviewer} \& \href{https://www.aclweb.org/adminwiki/index.php/ACL_Policy_on_Publication_Ethics\#Guidelines_for_Generative_Assistance_in_Authorship}{Author} guidelines.} challenges still remain. 
To address these, we propose the following based on our work and findings: 
(i) \textbf{Increase transparency in LLM usage within scholarly processes} by incorporating LLM assistance into review checklists, encouraging explicit acknowledgment of LLM support in paper acknowledgments, and 
reporting LLM usage patterns across diverse demographic groups;
% reporting LLM penetration based on social demographic features;
(ii) \textbf{Adopt policies to prevent irresponsible LLM reviewers} by establishing feedback channels for authors on LLM-generated reviews and developing fine-grained LLM detection models~\cite{abassy-etal-2024-llm, cheng2024beyond, artemova2025beemobenchmarkexperteditedmachinegenerated} to distinguish acceptable LLM roles (e.g., language improvement vs. content creation);
(iii) \textbf{Promote data-driven research in scholarly processes} by supporting the collection of review data for further robust analysis~\cite{dycke-etal-2022-yes}.\footnote{\url{https://arr-data.aclweb.org/}}

% make LLM usage transparent in scholarly processes: such as incorporating LLM usage into review checklists, encouraging explicit acknowledgment of LLM assistance in paper acknowledgments, and reporting LLM penetration based on social demographic features; (ii) Adopt policies to prevent irresponsible LLM reviewers: such as providing authors feedback on LLM-assisted reviews, and developing fine-grained LLM detection models~\cite{cheng2024beyond} to distinguish acceptable LLM roles (e.g., language improvement vs. content creation); (iii) Encourage data-driven research in scholarly processes: such as supporting review data collection for further research.

\section*{Limitations}
While this study provides valuable insights into the penetration of LLMs in scholarly writing and peer review, it may not fully represent the complexities of the real-world scenario. 
On one hand, the analysis focuses on peer review data from the ICLR conference, where the process is fully transparent and the quality of reviews is generally well-maintained. However, in many journals and conferences where peer review remains closed, the penetration of LLMs could be even more pronounced. 
On the other hand, the data simulation may not fully capture the intricate dynamics of LLM-human collaboration in real-world settings, making it difficult to distinguish between acceptable and unacceptable levels of LLM involvement, and potentially leading to a reduced ability of the model to detect LLM-generated text, which in turn lowers the assessment of LLM penetration.
Therefore, the penetration of large language models in scholarly writing and peer review may be more significant in real-world scenarios than what is presented in this study.
% \section*{Acknowledgments}

\bibliography{custom, anthology}

\appendix

\section{Prompts for Data Construction}
\label{app: prompts}

\begin{table*}[t]
\centering
\scalebox{0.85}{
\begin{tabular}{c|p{18cm}@{}}
\toprule
\textbf{Id} & \textbf{Prompt} \\  \midrule
1 & Can you help me revise the abstract? Please response directly with the revised abstract: \{abstract\} \\ 

2 & Please revise the abstract, and response directly with the revised abstract: \{abstract\} \\ 

3 & Can you check if the flow of the abstract makes sense? Please response directly with the revised abstract: \{abstract\} \\ 

4 & Please revise the abstract to make it more logical, response it directly with the revised abstract: \{abstract\} \\ 

5 & Please revise the abstract to make it more formal and academic, response it directly with the revised abstract: \{abstract\} \\ 
        \bottomrule

\end{tabular}}
\caption{Five distinct prompts used to refine human-written abstracts.}
\label{tab:prompt diversity}
\end{table*}

\begin{table*}[ht]
\centering
\scalebox{0.85}{
\begin{tabular}{@{}p{18.5cm}@{}}
\toprule

\multicolumn{1}{l}{\textbf{Basic Prompt Guideline}} \\ \cmidrule(r){1-1}
You are an AI assistant tasked with generating meta-reviews from multiple reviewers' feedback.\\
Please write a meta review of the given reviewers' response around \{$n$\} words.\\
Do not include any section titles or headings. 
Do not reference individual reviewers by name or number. 
Instead, focus on synthesizing collective feedback and overall opinion.

\\
\#\#\# Abstract: \{abstract\}

\#\#\# Reviewers' feedback:\{review\_text\}
\\ \midrule

\multicolumn{1}{l}{\textbf{Formatted Prompt Guideline 1}} \\ \cmidrule(r){1-1}
You are an AI assistant tasked with generating meta-reviews from multiple reviewers' feedback.\\
Please write a meta review of the given reviewers' response around \{$n$\} words. \\
Do not include any section titles or headings. Do not reference individual reviewers by name or number. Instead, focus on synthesizing collective feedback and overall opinion.

\

Please include the given format in your meta review:

Give a concise summary here.\\
Strength: [List the strengths of the paper in points based on reviews.]\\
Weakness: [List the weaknesses of the paper in points based on reviews.]

\\

\#\#\# Abstract: \{abstract\} \\
\#\#\# Reviewers' feedback:\{review\_text\}
\\ \midrule

\multicolumn{1}{l}{\textbf{Formatted Prompt Guideline 2}} \\ \midrule
You are an AI assistant tasked with generating meta-reviews from multiple reviewers' feedback.

Please write a meta review of the given reviewers' response around \{$n$\} words.

Do not include any section titles or headings. Do not reference individual reviewers by name or number.
Instead, focus on synthesizing collective feedback and overall opinion.

\

Please include the given format in your meta review:

Give a concise summary here.\\
Pros: [List the strengths of the paper in points based on reviews.]\\
Cons: [List the weaknesses of the paper in points based on reviews.]

\\

\#\#\# Abstract: \{abstract\}
\#\#\# Reviewers' feedback:\{review\_text\}
\\ \bottomrule

\end{tabular}}
\caption{Three prompts as guidelines for constructing LLM-generated meta-reviews. Here, $n$ represents the approximate word length for the generated content.}
\label{tab:prompt meta}
\end{table*}

\begin{table*}[t]
    \centering

    \begin{tabular}{>{\centering\arraybackslash}m{5cm}>{\centering\arraybackslash}m{3cm}>{\centering\arraybackslash}m{3cm}>{\centering\arraybackslash}m{3cm}}
        \toprule
        Word Count  & Basic Prompt & Formatted 1 & Formatted 2 \\
        \midrule
        $n \leq 50$ & 1.000 & 0.000 & 0.000 \\
        $50 < n \leq 110$ & 0.800 & 0.100 & 0.100 \\
        $110 < n \leq 160$ & 0.400 & 0.300 & 0.300 \\
        $160 < n \leq 220$ & 0.550 & 0.225 & 0.225 \\
        $> 220$ & 0.250 & 0.375 & 0.375 \\
        \bottomrule
    \end{tabular}
    \caption{The statistical distribution of word lengths observed for each involved meta-review format.}
    \label{tab:prompt frequency}
\end{table*}

\subsection{Prompts for abstract}
To ensure diversity in the refinement process, we design five different prompts for polishing the abstract, as shown in Table \ref{tab:prompt diversity}. Each human-written abstract is randomly assigned one prompt to generate the refined content.

\subsection{Prompts for meta-review}
To ensure that LLM-generated meta-reviews closely mirror the writing style of human-written meta-reviews and maintain authenticity, we analyze the characteristics of human-written meta-reviews. Based on this analysis, we provide three generation templates as guidelines for constructing LLM-generated meta-reviews, as shown in Table~\ref{tab:prompt meta}. The basic prompt does not include any formalized structures, while the other two prompts define more distinct meta-review formats.
Specifically, we conduct a detailed statistical analysis of the frequencies of these two paradigms and selected the corresponding prompts based on these frequencies. The probabilities we use are shown in Table \ref{tab:prompt frequency}.

\section{Dataset Details}
\label{app:dataset}
Table~\ref{tab: AcademicLens-statistics} shows the statistics of \texttt{ScholarLens}. 
The LLM versions used for data construction are: GPT-4o (\texttt{gpt-4o-2024-08-06}), Gemini-1.5 (\texttt{gemini-1.5-pro-002}), and Claude-3-Opus (\texttt{claude-3-opus-20240229}).

% GPT-4o (\texttt{gpt-4o-2024-08-06}), Gemini-1.5 (\texttt{gemini-1.5-pro-002}), Claude-3-Opus (\texttt{claude-3-opus-20240229}).

\begin{table*}[]
\scalebox{0.85}{
\begin{tabular}{@{}l|r|r|r@{}}
\toprule
\textbf{Academic Aspects} & \textbf{Human (Size)} & \textbf{LLM Source(Size)}                                & \textbf{Data Source}  \\ \midrule
\textbf{Abstract}         & \multirow{2}{*}{2831} & GPT-4o (2831) / Gemini-1.5 (2831) / Claude-3 Opus (2831) & \multirow{2}{*}{Ours} \\ \cmidrule(r){1-1} \cmidrule(lr){3-3}
\textbf{Meta-Review}      &                       & GPT-4o (2831) / Gemini-1.5 (2831) / Claude-3 Opus (2831) &                       \\ \midrule
\textbf{Review}           & 20 × |R|              & GPT-4 (20) / Gemini-1.5 (20) / Claude-3 Opus (20)        & ReviewCritique        \\ \bottomrule
\end{tabular}}
\caption{Statistics of \texttt{ScholarLens}: 20 × |R| represents 20 papers, each with |R| reviews, where |R| varies by paper.}
\label{tab: AcademicLens-statistics}
\end{table*}

% \usepackage{booktabs}
% \usepackage{multirow}
\begin{table}[]
\centering
\begin{tabular}{@{}l|l|c|r@{}}
\toprule
\textbf{Data Type}                    & \textbf{Data Source} & \textbf{Train}        & \textbf{Test} \\ \midrule
\multirow{2}{*}{\textbf{Abstract}}    & Human                & \multirow{4}{*}{1981} & 850           \\
                                      & LLM                  &                       & 2550          \\ \cmidrule(r){1-2} \cmidrule(l){4-4} 
\multirow{2}{*}{\textbf{Meta-Review}} & Human                &                       & 850           \\
                                      & LLM                  &                       & 2550          \\ \midrule
\multirow{2}{*}{\textbf{Review}}      & Human                & \multirow{2}{*}{-}    & 20            \\
                                      & LLM                  &                       & 60            \\ \bottomrule
\end{tabular}
\caption{Dataset split for training detection models.}
\label{tab:data_split}
\end{table}

\section{General Linguistic Metrics Implementation}
\label{app:general}
% We have 10 general linguistic metric: Average Word Length (AWL), Long Word Ratio (LWR), Stopword Ratio (SWR), Type Token Ratio (TTR), Average Sentence Length (ASL), Dependency Relation Variety (DRV), Subordinate Clause Density (SCD), Flesch Reading Ease (FRE), Sentiment Polarity Score (PS), and Sentiment Subjectivity Score (SS).
For word-level metrics, \texttt{NLTK} tokenization is used, and only alphabetic words are considered.
For sentence-level metrics, \texttt{spaCy} is used to process the text and extract features such as sentence length and the dependency relation label of each word. 
Additionally, Sentiment Polarity Score (PS) and Sentiment Subjectivity Score (SS) are evaluated using \texttt{TextBlob}, while FRE is calculated using \texttt{Textstat}.

% \lz{wait}

\section{Fine-Grained Detection Model}
\label{app:Fine-Grained}
Using the existing meta-review data from \texttt{ScholarLens} (including both human-written and LLM-synthesized versions), we apply the LLM-refined abstract construction method to generate an LLM-refined version for each human-written meta-review. We utilize GPT-4o as the single LLM source and train a fine-grained three-class detector on the meta-reviews using the same data split. This trained detection model is then used to predict the 2024 ICLR meta-reviews, with approximately 35.32\% predicted as LLM-refined and 1.39\% as LLM-synthesized, which  show that the LLM-refined role plays a more dominant part in LLM penetration.

\section{Case Study Details}
\label{app: cs-details}
\subsection{Word-Level Algorithm and Experiments}
\subsubsection{Hypothesis Testing Algorithm}
\label{app:TestingAlgorithm}
Building on the Two-Sample t-test, we propose a word-proportion-based method to identify word-level LLM preferences.
Specifically, given a set of pairs of human-written and LLM-generated texts $\mathcal{D} =\left\{ \left( x_{i}^{h}, x_{i}^{l} \right) \right\} $, where where $x_i^{h}$ represents the human-written case and $x_i^{l}$ represents the corresponding LLM-generated version, our goal is to determine whether a word $w$ is preferentially generated by the LLM.

\paragraph{(i) Word Proportion} 
We define the proportion of word $w$ appearing in the human-written set $\left\{ x_{i}^{h} \right\} $ and the LLM-generated set $\left\{ x_{i}^{l} \right\} $ as $\hat{p}_h\left( w \right) $ and $\hat{p}_l\left( w \right) $, respectively, representing the fraction of texts in which $w$ occurs:
\begin{equation}
    \hat{p}_h\left(w\right) ={\small{\frac{\mathrm{cnt}_h\left( w \right) +\epsilon}{\left| \mathcal{D} \right|}}}
\end{equation}
\begin{equation}
    \hat{p}_l(w) ={\small{\frac{\mathrm{cnt}_h\left( w \right) +\epsilon}{\left| \mathcal{D} \right|}}}
\end{equation}
where $\mathrm{cnt}\left( w \right) =\sum\nolimits_i^{}{\mathbb{I} \left( w\in x_{i}^{} \right)}$ counts the number of texts in the set ${x_i}$ where the word $x$ appears, with $\mathbb{I}(w \in x_i)$ being an indicator function that returns 1 if $w$ appears in $x_i$, and $\epsilon=1$ as a smoothing constant to account for words that do not appear in a given text.

\paragraph{(ii) Hypothesis Setting}
Then, we define the following two hypotheses:
\begin{itemize}
    \item \textbf{Null hypothesis} ($H_0$): $\hat{p}_h\left( w \right) \geqslant \hat{p}_l\left( w \right) $, suggesting that LLMs do \textbf{not} preferentially generate the word $w$.
    % suggests that the word $x$ appears more frequently, or equally, in human-written text compared to LLM-generated text, suggesting that LLMs do not preferentially generate the word $x$.
    % This implies that the proportion of the word's occurrence in LLM-generated texts is not higher than that in human-written texts.(i.e., the word is preferred by LLMs)
    \item \textbf{Alternative hypothesis} ($H_1$): $\hat{p}_h\left( w \right) < \hat{p}_l\left( w \right) $, suggesting that LLMs preferentially generate the word $w$.
    % which suggests that the word $x$ appears more frequently in LLM-generated text than in human-written text, implying a preference by LLMs.
    
    % This indicates that the proportion of the word's occurrence in LLM-generated texts is significantly higher than that in human-written texts.(i.e., the word is preferred by LLMs)
\end{itemize}

% Here, $\hat{p}_1$ and $\hat{p}_2$ represent the proportions of the word's occurrence in human-written texts and LLM-generated texts.

\paragraph{(iii) Hypothesis Testing}

Considering that the variance of word proportion may differ between the two text groups, we adopt Welch's t-test to quantify the difference. Specifically, the test statistic and degrees of freedom are computed as follows:

\begin{equation}
t(w)=\frac{\hat{p}_l\left( w \right) -\hat{p}_h\left( w \right)}{\sqrt{\small{\frac{s_{h}^{2}+s_{l}^{2}}{\left| \mathcal{D} \right|}}}}
\end{equation}
\begin{equation}
    df(w)=\frac{\left( \frac{s_{h}^{2}+s_{l}^{2}}{\left| \mathcal{D} \right|} \right) ^2}{\frac{(s_{h}^{2}/\left| \mathcal{D} \right|)^2+(s_{l}^{2}/\left| \mathcal{D} \right|)^2}{\left| \mathcal{D} \right|-1}}
\end{equation}

where $s_h$ and $s_l$ represent the standard deviations of the corresponding word proportions, calculated as follows:
\begin{equation}
    s=\sqrt{\frac{\hat{p}(1-\hat{p})}{\left| \mathcal{D} \right|}}
\end{equation}

% Here, $SE$ represents the standard error of the difference between the two proportions.

% \subparagraph{Standard Error Calculation}
% The standard error $SE_i$ of the sample proportion in the $i$-th type of text is:
% \begin{equation}
% SE_i=\sqrt{\frac{\hat{p}_i(1 - \hat{p}_i)}{n_i+\alphUI_{smooth}\cdot|V|}}
% \end{equation}
% The standard error $SE$ of the difference between the two proportions is:
% \begin{equation}
% SE=\sqrt{SE_1^2 + SE_2^2}
% \end{equation}

% \subparagraph{Degrees of Freedom Calculation}
% In general, for two independent samples with variances $s_1^2$ and $s_2^2$ and sample sizes $n_1$ and $n_2$, the original formula for the degrees of freedom $df$ in Welch's $t$-test is:
% \begin{equation}
% df = \frac{\left(\frac{s_1^2}{n_1}+\frac{s_2^2}{n_2}\right)^2}{\frac{(s_1^2 / n_1)^2}{n_1 - 1}+\frac{(s_2^2 / n_2)^2}{n_2 - 1}}
% \end{equation}
% In our context, with $s_i^2 = SE_i^2$ and considering the adjusted sample sizes $n_i'=n_i+\alphUI_{smooth}\cdot|V|$, the simplified formula for degrees of freedom is:
% \begin{equation}
% df=\frac{(SE_1^2 + SE_2^2)^2}{\frac{SE_1^4}{n_1+\alphUI_{smooth}\cdot|V| - 1}+\frac{SE_2^4}{n_2+\alphUI_{smooth}\cdot|V| - 1}}
% \end{equation}

\paragraph{(iv) Hypothesis Decision}
We define the critical t-value, $t_c$, as the threshold for rejecting or accepting the null hypothesis. 
It is calculated using the inverse of the cumulative distribution function (CDF) of the t-distribution:
\begin{equation}
    t_c(w) = t_{\alpha, \, df(w)}^{-1}
\end{equation}
where $\alpha=0.05$ is the significance level.
If $t(w) > t_c(w)$, we reject the null hypothesis and conclude that the word occurs significantly more often in LLM-generated texts than in human-written texts. In this way, we can identify the words favored by LLMs.

\subsubsection{Experimental Setup}

% \lz{@Ruijie, write: how we filter these words}
% 例如需要描写到的：考虑到一些词，在上下文可能以名词出现、也可能以动词出现，所以我们没有直接以word本身作为最小单位去filter，而是使用(word, pos)为最小单位……我们使用什么词性分析；我们不考虑停用词；我们不考虑大小写……我们使用什么作为排序的准则……

To address part-of-speech variability of the same word (e.g., `record' functioning as both a noun and a verb in a sentence), we adopt (word, POS) pairs as the fundamental unit for analysis rather than isolated words. We use SpaCy for POS tagging, and stopwords are excluded from consideration. 
Using the proposed Hypothesis Testing Algorithm (\S\ref{app:TestingAlgorithm}), we filter the LLM-preferred word set and rank these words based on their Word Usage Increase Ratio (WUIR), defined as follows:
\begin{equation}
    \mathrm{WUIR}\left(w\right) ={
    \small{\frac{\mathrm{cnt}_l\left( w \right) -\mathrm{cnt}_h\left(w\right)}{\mathrm{cnt}_h\left(w\right) +\epsilon}}}
\end{equation}

% \paragraph{(i) Preprocessing Rule}
% Before filtering, we implement three preprocessing rule on texts:
% \textbf{Stopword Removal}: Exclude tokens matching NLTK's English stopword list.
% \textbf{Case Insensitivity}: Convert all words to lowercase before processing.

% \paragraph{(ii) Word-POS Pair Unit}
% To address lexical ambiguity (e.g., words like "record" functioning as both nouns and verbs), we adopt (word, POS) pairs as the fundamental unit for analysis rather than isolated words. We use Spacy for POS tagging.

% \paragraph{(iii) Hypothesis Testing}
% For words that share the same POS(such as when focusing on adjectives (ADJ), we only consider words of this specific POS), we implement the Hypothesis Testing Algorithm. Through this process, we are able to identify the set of words that are preferred by LLMs.

% \paragraph{(iv) Ranking}
% The final word set is sorted by the metric 
% \begin{equation}
% \frac{\left( \sum_i \mathbb{I}(w \in x_i^l) + \epsilon \right) - \left( \sum_i \mathbb{I}(w \in x_i^h) + \epsilon \right)}{\sum_i \mathbb{I}(w \in x_i^h) + 1}
% \end{equation}
% in descending order.

\subsubsection{Results: LLM-Preferred words}
% \lz{@Ruijie, show the words result and briefly describe them }% 先展示GPT-4o的，分别简要描述在abstract和meta-review的倾好词，最后简要提及一下Claude和Gemini的结果，如表所示，不需要额外描述多的。
%For GPT4o, in meta-reviews, \textit{long words} account for 39.96\% and \textit{complex-syllabled words} account for 67.82\%. In abstracts, the proportions are 40.48\% and 73.45\% respectively.

%For GPT4o, in the meta - review, we obtained a set of preferred words consisting of 600 NOUN, 371 VERB, 275 ADJ, and 43 ADV.
%For its abstract, the set of preferred words we counted is composed of 201 NOUN, 422 VERB, 127 ADJ, and 65 ADV.
%The Top-30 LLM-preferred words of GPT4o in the meta-review and abstract are shown in Table~\ref{tab:4m}~\ref{tab:4a}.

\begin{table*}[t]
    \centering
    \resizebox{\textwidth}{!}{
        \begin{tabular}{@{}c|lrlrlr@{}}
            \toprule
            \textbf{POS} & \textbf{Word} & \textbf{WUIR} & \textbf{Word} & \textbf{WUIR} & \textbf{Word} & \textbf{WUIR}\\
            \midrule
            \multirow{10}{*}{\textbf{NOUN}} 
            & abstract & 40.00 & advancements & 28.50 & realm & 24.00 \\
            & alterations & 20.00 & aligns & 19.00 & methodologies & 18.80 \\
            & clarity & 17.00 & adaptability & 13.00 & surpasses & 13.00 \\
            & examination & 10.00 & competitiveness & 9.00 & aids & 9.00 \\
            & reliance & 8.75 & necessitating & 8.00 & assurances & 8.00 \\
            & necessitates & 8.00 & assertions & 8.00 & threats & 8.00 \\
            & assessments & 7.33 & advancement & 7.25 & enhancements & 7.25 \\
            & demands & 7.00 & findings & 6.91 & standpoint & 6.00 \\
            & oversight & 6.00 & study & 5.42 & exhibit & 5.33 \\
            & enhancement & 5.00 & adjustments & 5.00 & capitalizes & 5.00 \\
            \midrule
            \multirow{10}{*}{\textbf{VERB}} 
            & enhancing & 73.00 & necessitates & 58.00 & necessitating & 50.00 \\
            & featuring & 47.00 & revised & 46.00 & influenced & 40.00 \\
            & encompassing & 31.00 & enhances & 30.10 & showcasing & 29.00 \\
            & surpasses & 19.50 & underscoring & 19.00 & facilitating & 18.67 \\
            & necessitate & 18.00 & managing & 18.00 & concerning & 15.33 \\
            & garnered & 14.50 & employing & 14.06 & surpassing & 14.00 \\
            & adhere & 14.00 & neglecting & 14.00 & comprehend & 14.00 \\
            & underscore & 13.00 & discern & 13.00 & examines & 12.00 \\
            & accommodates & 11.00 & detail & 11.00 & utilizing & 10.87 \\
            & enhance & 10.82 & begins & 10.50 & integrating & 10.21 \\
            \midrule
            \multirow{10}{*}{\textbf{ADJ}} 
            & innovative & 34.17 & exceptional & 24.00 & pertinent & 22.00 \\
            & intricate & 16.33 & pivotal & 16.00 & necessitate & 12.00 \\
            & distinctive & 11.00 & enhanced & 10.80 & akin & 10.40 \\
            & potent & 10.00 & adaptable & 9.67 & unfamiliar & 9.00 \\
            & straightforward & 8.42 & accessible & 8.00 & versatile & 7.13 \\
            & adept & 7.00 & devoid & 7.00 & advantageous & 6.80 \\
            & extended & 6.67 & prevalent & 6.00 & underexplored & 6.00 \\
            & commendable & 6.00 & contingent & 6.00 & foundational & 5.75 \\
            & comprehensive & 5.09 & strategic & 5.00 & renowned & 5.00 \\
            & attributable & 5.00 & unidentified & 5.00 & numerous & 4.82 \\
            \midrule
            \multirow{10}{*}{\textbf{ADV}} 
            & inadequately & 17.00 & predominantly & 16.67 & meticulously & 16.00 \\
            & strategically & 14.00 & notably & 12.60 & abstract & 12.00 \\
            & swiftly & 12.00 & additionally & 9.08 & adeptly & 8.00 \\
            & thereby & 7.52 & conversely & 7.40 & traditionally & 7.33 \\
            & initially & 7.10 & innovatively & 7.00 & subsequently & 6.06 \\
            & unexpectedly & 6.00 & excessively & 5.00 & historically & 5.00 \\
            & seamlessly & 4.50 & nonetheless & 4.25 & primarily & 4.00 \\
            & markedly & 4.00 & short & 4.00 & infrequently & 4.00 \\
            & effectively & 3.97 & solely & 3.80 & consequently & 3.78 \\
            & inherently & 3.40 & concurrently & 3.33 & particularly & 3.09 \\

            \bottomrule
        \end{tabular}
    }
    \caption{Top-30 LLM-preferred Words in \textbf{GPT-4o}-generated vs. human-written \textbf{abstracts}, with long words making up 40.48\%, and complex-syllabled words 73.45\%.}
    \label{tab:4a}
\end{table*}
\begin{table*}[t]
    \centering
    \resizebox{\textwidth}{!}{
        \begin{tabular}{@{}c|lrlrlr@{}}
            \toprule
            \textbf{POS} & \textbf{Word} & \textbf{WUIR} & \textbf{Word} & \textbf{WUIR} & \textbf{Word} & \textbf{WUIR}\\
            \midrule
            \multirow{10}{*}{\textbf{NOUN}} 
            & refinement & 284.00 & advancements & 88.00 & methodologies & 58.50 \\
            & articulation & 50.00 & highlights & 41.00 & reliance & 39.50 \\
            & enhancement & 39.00 & underpinnings & 37.00 & enhancements & 28.33 \\
            & transparency & 26.00 & complexities & 25.00 & skepticism & 24.67 \\
            & adaptability & 24.00 & narrative & 24.00 & integration & 23.43 \\
            & persist & 22.00 & acknowledgment & 22.00 & differentiation & 21.67 \\
            & advancement & 21.33 & contextualization & 21.00 & foundation & 20.50 \\
            & inconsistencies & 20.00 & reception & 20.00 & demands & 20.00 \\
            & backing & 19.50 & sections & 18.71 & refinements & 17.50 \\
            & credibility & 16.50 & benchmarking & 16.00 & reliability & 15.50 \\
            \midrule
            \multirow{10}{*}{\textbf{VERB}} 
            & enhance & 239.00 & enhancing & 120.50 & deemed & 79.33 \\
            & showcasing & 66.00 & express & 52.17 & offering & 50.50 \\
            & enhances & 45.00 & recognizing & 41.00 & commend & 37.25 \\
            & praised & 37.20 & integrating & 36.25 & criticized & 32.00 \\
            & hindering & 32.00 & utilizes & 31.00 & highlights & 28.00 \\
            & surpass & 27.00 & bolster & 26.00 & emphasizing & 25.33 \\
            & substantiate & 25.00 & integrates & 24.50 & solidify & 23.00 \\
            & questioning & 22.67 & arise & 21.17 & expanding & 20.67 \\
            & faces & 20.67 & weakens & 20.00 & recognized & 19.71 \\
            & criticize & 19.00 & illustrating & 19.00 & critique & 19.00 \\
            \midrule
            \multirow{10}{*}{\textbf{ADJ}} 
            & innovative & 114.50 & collective & 99.00 & enhanced & 45.00 \\
            & established & 30.00 & notable & 28.00 & outdated & 22.00 \\
            & varied & 16.00 & undefined & 15.00 & comparative & 14.70 \\
            & noteworthy & 14.50 & broader & 14.10 & comprehensive & 13.67 \\
            & clearer & 13.61 & intriguing & 13.50 & foundational & 13.00 \\
            & organizational & 13.00 & typographical & 12.50 & contextual & 12.00 \\
            & traditional & 11.83 & advanced & 11.00 & inadequate & 10.89 \\
            & diverse & 9.84 & insightful & 9.56 & prevalent & 9.50 \\
            & spatiotemporal & 9.00 & engaging & 9.00 & adaptable & 9.00 \\
            & illustrative & 8.50 & robotic & 8.50 & commendable & 8.33 \\
            \midrule
            \multirow{10}{*}{\textbf{ADV}} 
            & collectively & 223.00 & inadequately & 38.00 & reportedly & 18.00 \\
            & comprehensively & 13.00 & robustly & 12.00 & occasionally & 10.00 \\
            & predominantly & 9.00 & notably & 8.38 & innovatively & 8.00 \\
            & effectively & 7.81 & insufficiently & 6.71 & additionally & 6.59 \\
            & particularly & 5.09 & creatively & 5.00 & distinctly & 5.00 \\
            & positively & 4.71 & overall & 4.30 & primarily & 3.32 \\
            & convincingly & 3.07 & elegantly & 3.00 & marginally & 2.93 \\
            & selectively & 2.50 & conclusively & 2.33 & especially & 2.26 \\
            & favorably & 2.20 & universally & 2.00 & theoretically & 1.71 \\
            & overly & 1.71 & consistently & 1.65 & potentially & 1.59 \\
            \bottomrule
        \end{tabular}
    }
    \caption{Top-30 LLM-preferred Words in \textbf{GPT-4o}-generated vs. human-written \textbf{meta-reviews}, with long words making up 39.96\% and complex-syllabled words 67.82\%.}
    \label{tab:4m}
\end{table*}

Tables~\ref{tab:4a} and~\ref{tab:4m} display the top-30 preferred words across four key part-of-speech (POS) categories in GPT-4o-generated abstracts and meta-reviews, with long words accounting for 40.48\% and 39.96\%, and complex-syllabled words for 73.45\% and 67.82\%. Furthermore, Tables~\ref{tab:ga} and~\ref{tab:gm} show the top-30 preferred words in four key POS categories for Gemini-generated abstracts and meta-reviews, while Tables~\ref{tab:ca} and~\ref{tab:cm} display the same for Claude-generated abstracts and meta-reviews. All show a high proportion of long words and complex-syllabled words.

\begin{table*}[t]
    \centering
    \resizebox{\textwidth}{!}{
        \begin{tabular}{@{}c|lrlrlr@{}}
            \toprule
            \textbf{POS} & \textbf{Word} & \textbf{WUIR} & \textbf{Word} & \textbf{WUIR} & \textbf{Word} & \textbf{WUIR}\\
            \midrule
            \multirow{10}{*}{\textbf{NOUN}} 
            & abstract & 40.00 & advancements & 28.50 & realm & 24.00 \\
            & alterations & 20.00 & aligns & 19.00 & methodologies & 18.80 \\
            & clarity & 17.00 & adaptability & 13.00 & surpasses & 13.00 \\
            & examination & 10.00 & competitiveness & 9.00 & aids & 9.00 \\
            & reliance & 8.75 & necessitating & 8.00 & assurances & 8.00 \\
            & necessitates & 8.00 & assertions & 8.00 & threats & 8.00 \\
            & assessments & 7.33 & advancement & 7.25 & enhancements & 7.25 \\
            & demands & 7.00 & findings & 6.91 & standpoint & 6.00 \\
            & oversight & 6.00 & study & 5.42 & exhibit & 5.33 \\
            & enhancement & 5.00 & adjustments & 5.00 & capitalizes & 5.00 \\
            \midrule
            \multirow{10}{*}{\textbf{VERB}} 
            & enhancing & 73.00 & necessitates & 58.00 & necessitating & 50.00 \\
            & featuring & 47.00 & revised & 46.00 & influenced & 40.00 \\
            & encompassing & 31.00 & enhances & 30.10 & showcasing & 29.00 \\
            & surpasses & 19.50 & underscoring & 19.00 & facilitating & 18.67 \\
            & necessitate & 18.00 & managing & 18.00 & concerning & 15.33 \\
            & garnered & 14.50 & employing & 14.06 & surpassing & 14.00 \\
            & adhere & 14.00 & neglecting & 14.00 & comprehend & 14.00 \\
            & underscore & 13.00 & discern & 13.00 & examines & 12.00 \\
            & accommodates & 11.00 & detail & 11.00 & utilizing & 10.87 \\
            & enhance & 10.82 & begins & 10.50 & integrating & 10.21 \\
            \midrule
            \multirow{10}{*}{\textbf{ADJ}} 
            & innovative & 34.17 & exceptional & 24.00 & pertinent & 22.00 \\
            & intricate & 16.33 & pivotal & 16.00 & necessitate & 12.00 \\
            & distinctive & 11.00 & enhanced & 10.80 & akin & 10.40 \\
            & potent & 10.00 & adaptable & 9.67 & unfamiliar & 9.00 \\
            & straightforward & 8.42 & accessible & 8.00 & versatile & 7.13 \\
            & adept & 7.00 & devoid & 7.00 & advantageous & 6.80 \\
            & extended & 6.67 & prevalent & 6.00 & underexplored & 6.00 \\
            & commendable & 6.00 & contingent & 6.00 & foundational & 5.75 \\
            & comprehensive & 5.09 & strategic & 5.00 & renowned & 5.00 \\
            & attributable & 5.00 & unidentified & 5.00 & numerous & 4.82 \\
            \midrule
            \multirow{10}{*}{\textbf{ADV}} 
            & inadequately & 17.00 & predominantly & 16.67 & meticulously & 16.00 \\
            & strategically & 14.00 & notably & 12.60 & abstract & 12.00 \\
            & swiftly & 12.00 & additionally & 9.08 & adeptly & 8.00 \\
            & thereby & 7.52 & conversely & 7.40 & traditionally & 7.33 \\
            & initially & 7.10 & innovatively & 7.00 & subsequently & 6.06 \\
            & unexpectedly & 6.00 & excessively & 5.00 & historically & 5.00 \\
            & seamlessly & 4.50 & nonetheless & 4.25 & primarily & 4.00 \\
            & markedly & 4.00 & short & 4.00 & infrequently & 4.00 \\
            & effectively & 3.97 & solely & 3.80 & consequently & 3.78 \\
            & inherently & 3.40 & concurrently & 3.33 & particularly & 3.09 \\
            \bottomrule
        \end{tabular}
    }
    \caption{Top-30 LLM-preferred Words in \textbf{Gemini}-generated vs. human-written \textbf{abstracts}, with long words making up 44.35\%, and complex-syllabled words 74.80\%}
    \label{tab:ga}
\end{table*}
\begin{table*}[t]
    \centering
    \resizebox{\textwidth}{!}{
        \begin{tabular}{@{}c|lrlrlr@{}}
            \toprule
            \textbf{POS} & \textbf{Word} & \textbf{WUIR} & \textbf{Word} & \textbf{WUIR} & \textbf{Word} & \textbf{WUIR}\\
            \midrule
            \multirow{10}{*}{\textbf{NOUN}} 
            & refinement & 59.00 & leans & 55.00 & reliance & 45.50 \\
            & generalizability & 37.00 & handling & 35.00 & advancements & 28.00 \\
            & achieves & 24.00 & explores & 23.00 & inconsistencies & 20.33 \\
            & underpinnings & 17.00 & core & 16.38 & clarification & 16.29 \\
            & hinder & 16.00 & contingent & 16.00 & implications & 14.42 \\
            & articulation & 14.00 & calculations & 14.00 & typos & 13.86 \\
            & quantification & 13.00 & testing & 12.71 & availability & 12.50 \\
            & efficacy & 12.33 & referencing & 12.00 & mitigation & 12.00 \\
            & contextualization & 12.00 & duration & 12.00 & investigation & 11.97 \\
            & practicality & 11.43 & grounding & 11.25 & overfitting & 11.00 \\
            \midrule
            \multirow{10}{*}{\textbf{VERB}} 
            & deemed & 153.67 & solidify & 94.00 & hindering & 66.00 \\
            & criticized & 63.67 & praised & 52.20 & weakens & 42.00 \\
            & exceeding & 41.00 & praising & 40.33 & questioned & 37.68 \\
            & leans & 32.57 & recognizing & 32.00 & drew & 31.00 \\
            & leaned & 29.50 & enhance & 28.00 & offering & 27.25 \\
            & mitigating & 27.00 & weakened & 26.50 & utilizes & 25.00 \\
            & showcasing & 23.00 & hinders & 23.00 & arose & 21.67 \\
            & expanding & 20.67 & recurring & 19.00 & challenged & 17.00 \\
            & leverages & 16.75 & desired & 15.46 & raising & 15.33 \\
            & promoting & 15.00 & termed & 15.00 & differing & 15.00 \\
            \midrule
            \multirow{10}{*}{\textbf{ADJ}} 
            & core & 65.25 & established & 29.00 & undefined & 17.00 \\
            & nuanced & 16.00 & cautious & 16.00 & presentational & 15.00 \\
            & absent & 13.00 & combined & 12.00 & innovative & 11.62 \\
            & outdated & 11.00 & simplified & 9.25 & robotic & 9.00 \\
            & benchmark & 8.56 & inconsistent & 8.50 & illustrative & 8.50 \\
            & diverse & 8.26 & adaptable & 8.00 & dataset & 7.44 \\
            & insightful & 7.33 & grammatical & 7.29 & observed & 7.20 \\
            & comprehensive & 7.03 & spatiotemporal & 7.00 & repetitive & 7.00 \\
            & certain & 6.98 & rigorous & 6.86 & deeper & 6.64 \\
            & superior & 6.55 & compact & 6.50 & unconvincing & 6.30 \\
            \midrule
            \multirow{10}{*}{\textbf{ADV}} 
            & definitively & 9.00 & solely & 8.43 & reportedly & 8.00 \\
            & particularly & 7.35 & purportedly & 7.00 & primarily & 6.89 \\
            & positively & 5.43 & generally & 5.02 & straightforward & 5.00 \\
            & adaptively & 5.00 & specifically & 4.96 & favorably & 4.80 \\
            & demonstrably & 4.00 & locally & 3.50 & consistently & 3.41 \\
            & incrementally & 3.25 & potentially & 3.21 & visually & 3.00 \\
            & furthermore & 2.93 & theoretically & 2.58 & overhead & 2.50 \\
            & effectively & 2.35 & publicly & 2.29 & additionally & 2.15 \\
            & fine & 1.88 & especially & 1.76 & computationally & 1.71 \\
            & finally & 1.62 & insufficiently & 1.50 & overly & 1.47 \\
            \bottomrule
        \end{tabular}
    }
    \caption{Top-30 LLM-preferred Words in \textbf{Gemini}-generated vs. human-written \textbf{meta-reviews}, with long words making up 34.58\%, and complex-syllabled words 62.07\%}
    \label{tab:gm}
\end{table*}

\begin{table*}[t]
    \centering
    \resizebox{\textwidth}{!}{
        \begin{tabular}{@{}c|lrlrlr@{}}
            \toprule
            \textbf{POS} & \textbf{Word} & \textbf{WUIR} & \textbf{Word} & \textbf{WUIR} & \textbf{Word} & \textbf{WUIR}\\
            \midrule
            \multirow{10}{*}{\textbf{NOUN}} 
            & abstract & 40.00 & advancements & 28.50 & realm & 24.00 \\
            & alterations & 20.00 & aligns & 19.00 & methodologies & 18.80 \\
            & clarity & 17.00 & adaptability & 13.00 & surpasses & 13.00 \\
            & examination & 10.00 & competitiveness & 9.00 & aids & 9.00 \\
            & reliance & 8.75 & necessitating & 8.00 & assurances & 8.00 \\
            & necessitates & 8.00 & assertions & 8.00 & threats & 8.00 \\
            & assessments & 7.33 & advancement & 7.25 & enhancements & 7.25 \\
            & demands & 7.00 & findings & 6.91 & standpoint & 6.00 \\
            & oversight & 6.00 & study & 5.42 & exhibit & 5.33 \\
            & enhancement & 5.00 & adjustments & 5.00 & capitalizes & 5.00 \\
            \midrule
            \multirow{10}{*}{\textbf{VERB}} 
            & enhancing & 73.00 & necessitates & 58.00 & necessitating & 50.00 \\
            & featuring & 47.00 & revised & 46.00 & influenced & 40.00 \\
            & encompassing & 31.00 & enhances & 30.10 & showcasing & 29.00 \\
            & surpasses & 19.50 & underscoring & 19.00 & facilitating & 18.67 \\
            & necessitate & 18.00 & managing & 18.00 & concerning & 15.33 \\
            & garnered & 14.50 & employing & 14.06 & surpassing & 14.00 \\
            & adhere & 14.00 & neglecting & 14.00 & comprehend & 14.00 \\
            & underscore & 13.00 & discern & 13.00 & examines & 12.00 \\
            & accommodates & 11.00 & detail & 11.00 & utilizing & 10.87 \\
            & enhance & 10.82 & begins & 10.50 & integrating & 10.21 \\
            \midrule
            \multirow{10}{*}{\textbf{ADJ}} 
            & innovative & 34.17 & exceptional & 24.00 & pertinent & 22.00 \\
            & intricate & 16.33 & pivotal & 16.00 & necessitate & 12.00 \\
            & distinctive & 11.00 & enhanced & 10.80 & akin & 10.40 \\
            & potent & 10.00 & adaptable & 9.67 & unfamiliar & 9.00 \\
            & straightforward & 8.42 & accessible & 8.00 & versatile & 7.13 \\
            & adept & 7.00 & devoid & 7.00 & advantageous & 6.80 \\
            & extended & 6.67 & prevalent & 6.00 & underexplored & 6.00 \\
            & commendable & 6.00 & contingent & 6.00 & foundational & 5.75 \\
            & comprehensive & 5.09 & strategic & 5.00 & renowned & 5.00 \\
            & attributable & 5.00 & unidentified & 5.00 & numerous & 4.82 \\
            \midrule
            \multirow{10}{*}{\textbf{ADV}} 
            & inadequately & 17.00 & predominantly & 16.67 & meticulously & 16.00 \\
            & strategically & 14.00 & notably & 12.60 & abstract & 12.00 \\
            & swiftly & 12.00 & additionally & 9.08 & adeptly & 8.00 \\
            & thereby & 7.52 & conversely & 7.40 & traditionally & 7.33 \\
            & initially & 7.10 & innovatively & 7.00 & subsequently & 6.06 \\
            & unexpectedly & 6.00 & excessively & 5.00 & historically & 5.00 \\
            & seamlessly & 4.50 & nonetheless & 4.25 & primarily & 4.00 \\
            & markedly & 4.00 & short & 4.00 & infrequently & 4.00 \\
            & effectively & 3.97 & solely & 3.80 & consequently & 3.78 \\
            & inherently & 3.40 & concurrently & 3.33 & particularly & 3.09 \\
            \bottomrule
        \end{tabular}
    }
    \caption{Top-30 LLM-preferred Words in \textbf{Claude}-generated vs. human-written \textbf{abstracts}, with long words making up 42.81\%, and complex-syllabled words 74.80\%}
    \label{tab:ca}
\end{table*}

\begin{table*}[t]
    \centering
    \resizebox{\textwidth}{!}{
        \begin{tabular}{@{}c|lrlrlr@{}}
            \toprule
            \textbf{POS} & \textbf{Word} & \textbf{WUIR} & \textbf{Word} & \textbf{WUIR} & \textbf{Word} & \textbf{WUIR}\\
            \midrule
            \multirow{10}{*}{\textbf{NOUN}} 
            & refinement & 604.00 & center & 151.00 & demonstrates & 108.00 \\
            & foundations & 99.00 & articulation & 70.00 & relies & 62.00 \\
            & generalizability & 60.57 & tier & 38.00 & reservations & 37.43 \\
            & sentiments & 37.00 & critiques & 34.00 & explores & 31.00 \\
            & vulnerabilities & 29.00 & transparency & 27.00 & achieves & 27.00 \\
            & promise & 25.67 & advancement & 25.67 & substantiation & 25.00 \\
            & introduces & 24.00 & benchmarking & 23.00 & leans & 23.00 \\
            & underpinnings & 22.00 & capabilities & 21.00 & skepticism & 21.00 \\
            & shows & 20.00 & challenges & 19.53 & ambiguities & 19.00 \\
            & narrative & 19.00 & highlights & 19.00 & differentiation & 18.67 \\
            \midrule
            \multirow{10}{*}{\textbf{VERB}} 
            & synthesizes & 984.00 & recognizing & 153.00 & view & 131.33 \\
            & critique & 92.00 & express & 91.83 & revealing & 81.33 \\
            & appreciating & 73.67 & praising & 56.67 & expanding & 52.67 \\
            & highlighting & 47.53 & offering & 42.50 & acknowledging & 39.92 \\
            & center & 37.50 & mitigating & 36.00 & substantiate & 32.67 \\
            & persist & 27.00 & enhance & 26.50 & deemed & 26.00 \\
            & conducting & 24.00 & reveals & 24.00 & graph & 23.00 \\
            & handling & 20.67 & contexts & 20.00 & emerge & 19.75 \\
            & recognize & 19.43  & generative & 17.50 
            & represents & 17.10\\ & praised & 16.80 & offers & 16.66 & bridging&16.00 \\
            \midrule
            \multirow{10}{*}{\textbf{ADJ}} 
            & collective & 1049.00 & nuanced & 292.00 & innovative & 109.50 \\
            & cautious & 46.00 & comprehensive & 40.22 & comparative & 36.30 \\
            & revolutionary & 33.00 & definitive & 31.00 & meta & 27.84 \\
            & methodological & 22.28 & rigorous & 22.24 & transformative & 21.00 \\
            & noteworthy & 18.50 & substantive & 18.00 & presentational & 18.00 \\
            & notable & 17.75 & clearer & 17.06 & scholarly & 17.00 \\
            & academic & 16.00 & undefined & 15.00 
            & intriguing & 14.00\\ & addresses & 14.00 & robotic & 14.00 
            & core & 13.38 \\& diverse & 12.11 & adaptable & 12.00
            & primary & 11.26\\ & scientific & 10.69 & deeper & 10.64 &broader&10.17\\
            \midrule
            \multirow{10}{*}{\textbf{ADV}} 
            & collectively & 1112.00 & definitively & 54.00 & comprehensively & 37.00 \\
            & positively & 25.14 & conclusively & 16.00 & critically & 16.00 \\
            & consistently & 15.24 & scientifically & 12.00 & marginally & 8.57 \\
            & predominantly & 7.00 & cautiously & 6.50 & unanimously & 5.94 \\
            & robustly & 5.00 & adaptively & 5.00 & particularly & 4.59 \\
            & incrementally & 4.50 & meaningfully & 4.33 & generally & 4.11 \\
            & fully & 3.95 & primarily & 3.61 & overhead & 3.50 \\
            & short & 3.38 & potentially & 3.37 & genuinely & 3.33 \\
            & dynamically & 2.78 & fundamentally & 2.71 & technically & 2.54 \\
            & methodologically & 2.50 & semantically & 2.33 & dramatically & 2.33 \\
            \bottomrule
        \end{tabular}
    }
    \caption{Top-30 LLM-preferred Words in \textbf{Claude}-generated vs. human-written \textbf{meta-reviews}, with long words making up 43.93\%, and complex-syllabled words 70.49\%}
    \label{tab:cm}
\end{table*}

\subsection{Pattern-Level Feature Statistics}
We identify the following pattern-level features in human-written (meta-)reviews: \textit{personability}, characterized by frequent use of the first person to express opinions; \textit{interactivity}, marked by the inclusion of questions; and \textit{attention to detail}, demonstrated by citing relevant literature to support arguments. 
To compare these pattern-level features between human-written and LLM-generated content, we calculate two metrics for each pattern in both meta-reviews and reviews within the ScholarLens dataset: \textbf{Feature Proportion (FP)} and \textbf{Feature Intensity (FI)}. FP is defined as the proportion of instances exhibiting the feature within the target data group, while FI is the average number of occurrences of the feature within instances that exhibit it. We report the FP and FI values for each pattern in meta-reviews and reviews across different data types—Human-written, GPT-4-generated, Gemini-generated, and Claude-generated— as shown in Table~\ref{tab:pattern_comparison}.
% (\textit{Personability} Pattern Comparison), Table~\ref{tab:Interactivity_c} (\textit{Interactivity} Pattern Comparison), and Table~\ref{tab:Attention_c} (\textit{Attention to Detail} Pattern Comparison).
The results show that, in both meta-reviews and reviews, the FP and FI values for each pattern feature in the Human-written data type are significantly higher than those in the LLM-generated versions.

\subsection{Validation of Detection Model Reliability}
We use our filtered full LLM-preferred word set and the identified pattern-level features to validate the reliability of our detection models.
Specifically, we classify all meta-reviews from ICLR 2024 into two groups based on the fine-grained detection results in Appendix~\S\ref{app:Fine-Grained}: human-written and LLM-generated (including LLM-refined and LLM-synthesized prediction). For each meta-review, we calculate the proportion of words that belong to the full GPT-4-preferred word set, defined as the ratio of matching words to the total number of words in the set. The average ratios for the human-written and LLM-generated groups are 35.15\% and 44.95\%, respectively. 
We then compute the FR and FI values for each pattern feature in each group, with results shown in Table~\ref{tab:pattern_vali}. The FR and FI values for predicted LLM-generated text are lower than those for predicted human-written text.
% These results  demonstrate the reliability of our model's detection capabilities.
These results provide evidence of the reliability of our model's detection capabilities.
% Then we compute the FR and FI value of each pattern feature in each group, the results are shown in Table~\ref{tab:pattern_vali}, 预测为LLM-generated text呈现的FR和FI value 低于预测为human-written的text

% \clearpage

\begin{figure}[h]
    \centering
    \begin{minipage}{1\linewidth}
        \centering
        \scalebox{0.95}{
        \begin{tabular}{l|l|r|r}
        \toprule
        {\textbf{Data Type}}&\textbf{Resource }&\textbf{FR (\%)}&\textbf{FI}\\
        \midrule
        \multirow{5}{*}{Meta-review} 

            & Human & 32.00& 1.62\\
            & GPT4o & 0.07 & 1.00\\
            & Gemini & 0.11& 1.33\\
            & Claude & 0.07& 1.00\\
        \midrule
        \multirow{4}{*}{Review} 
            & Human & 76.32 & 2.07\\
            & GPT4o &0.00&0.00\\
            & Gemini &0.00&0.00\\
            & Claude & 0.00&0.00\\
        \bottomrule
    \end{tabular}}
        % \captionsetup{font=footnotesize}
        \subcaption{\textit{Personability}}
    \end{minipage}
    \hfill
    \begin{minipage}{1\linewidth}
        \centering
        \scalebox{0.95}{
        \begin{tabular}{l|l|r|r}
        \hline
        {\textbf{Data Type}}&\textbf{Resource }&\textbf{FR (\%)}&\textbf{FI}\\
        \toprule
        \multirow{5}{*}{Meta-review} 

            & Human& 2.01& 1.54 \\
            & GPT4o& 0.00&0.00 \\
            & Gemini& 0.00&0.00 \\
            & Claude& 0.00&0.00 \\
        \midrule
        \multirow{4}{*}{Review} 
            & Human& 17.11& 1.85 \\
            & GPT4o& 0.00&0.00 \\
            & Gemini& 0.00&0.00 \\
            & Claude& 0.00&0.00 \\
        \bottomrule
    \end{tabular}}
        % \captionsetup{font=footnotesize}
        \subcaption{\textit{Interactivity}}
    \end{minipage}
    \hfill
    \begin{minipage}{1\linewidth}
        \centering
        \scalebox{0.95}{
        \begin{tabular}{l|l|r|r}
        \toprule
        {\textbf{Data Type}}&\textbf{Resource }&\textbf{FR (\%)}&\textbf{FI}\\
        \midrule
        \multirow{5}{*}{Meta-review} 
            & Human & 1.48&1.60\\
            & GPT4o & 0.00&0.00\\
            & Gemini & 0.00&0.00\\
            & Claude & 0.00&0.00\\
%            & meta\_overall & 0/519\\
%            & meta\_overall\_except & 392/17003 \\
        \bottomrule
        \multirow{4}{*}{Review} 
            & Human & 7.89&3.17\\
            & GPT4o & 0.00&0.00\\
            & Gemini & 0.00&0.00\\
            & Claude & 0.00&0.00\\
        \hline
    \end{tabular}}
        % \captionsetup{font=footnotesize}
        \subcaption{\textit{Attention to Detail}}
    \end{minipage}
    \hfill

    \caption{Comparison of Pattern Features of Meta-Reviews and Reviews in \texttt{ScholarLens}.}
    \label{tab:pattern_comparison}
\end{figure}

\begin{figure}[ht]
    \centering
    \begin{minipage}{1\linewidth}
        \centering
        \begin{tabular}{l|l|r|r}
        \toprule
\textbf{Role} &\textbf{FR (\%)}&\textbf{FI}\\
        \midrule

LLM-synthesized&6.12&1.00\\
LLM-refined&34.32&1.53\\
LLM-generated&32.61&1.53\\
human-written &39.50&1.77\\
%            & meta\_meta & 10.77&1.18\\
%            & meta\_meta\_comp & 40.11&1.72 \\
        \bottomrule
    \end{tabular}
        % \captionsetup{font=footnotesize}
        \subcaption{\textit{Personability}}
    \end{minipage}
    \hfill
    \begin{minipage}{1\linewidth}
        \centering
        \begin{tabular}{l|r|r}
        \toprule
\textbf{Role} & \textbf{FR (\%)}&\textbf{FI}\\
        \midrule

LLM-synthesized &0.00&0.00\\
LLM-refined &2.02 &1.53\\
LLM-generated &1.89 &1.53\\
human-written &4.87 &1.88\\
%            & meta\_meta & 2/520&1.00\\
%            & meta\_meta\_comp & 229/5260&1.84 \\
        \bottomrule
    \end{tabular}
        % \captionsetup{font=footnotesize}
        \subcaption{\textit{Interactivity}}
    \end{minipage}
    \hfill
    \begin{minipage}{1\linewidth}
        \centering
        \begin{tabular}{l|r|r}
        \toprule
        \textbf{Role} & \textbf{FR (\%)}&\textbf{FI}\\
        \midrule
            LLM-synthesized &0.00&0.00\\
            LLM-refined &1.57&1.13\\
            LLM-generated &1.47&1.06\\
            human-written &3.77&2.25\\
%            & meta\_meta & 1/520\\
%            & meta\_meta\_comp & 109/5260 \\
        \bottomrule
    \end{tabular}
        % \captionsetup{font=footnotesize}
        \subcaption{\textit{Attention to Detail}}
    \end{minipage}
    \hfill

    \caption{Comparison of Pattern Features in Each Prediction Data Group for ICLR 2024 Meta-Reviews.}
    \label{tab:pattern_vali}
\end{figure}

\end{document}